\documentclass[11pt]{article} % arxiv

\let\citep\cite %arxiv
\let\citet\cite % arxiv

\PassOptionsToPackage{sort,numbers}{natbib}

\usepackage[final]{neurips_2024}

\usepackage[utf8]{inputenc} % allow utf-8 input
\usepackage[T1]{fontenc}    % use 8-bit T1 fonts
\usepackage{url}            % simple URL typesetting
\usepackage{amsfonts}       % blackboard math symbols
\usepackage{nicefrac}       % compact symbols for 1/2, etc.
\usepackage{microtype}      % microtypography
\usepackage{microtype}
\usepackage{graphicx}
\usepackage{subfigure}
\usepackage{booktabs} % for professional tables
\usepackage[table,xcdraw]{xcolor}
\usepackage[colorlinks=true, linkcolor=red, urlcolor=blue, citecolor=blue]{hyperref}

\usepackage{wrapfig}

\usepackage{algorithm}
\usepackage{algorithmic}

\usepackage{amsmath}
\usepackage{amssymb}
\usepackage{mathtools}
\usepackage{amsthm}
\usepackage{enumitem}
\usepackage{comment}

\usepackage{pifont}
\newcommand{\cmark}{\ding{51}}%
\usepackage[nameinlink,capitalise,noabbrev]{cleveref}

\usepackage{xcolor}
\definecolor{dark-blue}{rgb}{0.15,0.15,0.4}
\hypersetup{
    colorlinks=true,
    linkcolor=blue,
    filecolor=magenta,      
    urlcolor=cyan,
    citecolor=dark-blue,
    pdftitle={TuneTables: Context Optimization for Scalable Prior-Data Fitted Networks},
    pdfpagemode=FullScreen,
}

\usepackage[textsize=tiny]{todonotes}

\theoremstyle{plain}

\theoremstyle{definition}

\theoremstyle{remark}

\DeclareMathOperator*{\argmin}{arg\,min}

\newcommand{\R}{\mathbb{R}}%
\newcommand{\E}{\mathbb{E}}

\newcommand{\ndatasets}{98}%
\newcommand{\nregdatasets}{15}%

\newcommand{\newbenchmark}{\textsc{LargeScaleTables}~}

\title{TuneTables: Context Optimization for Scalable \\ Prior-Data Fitted Networks}

\author{Benjamin Feuer$^{1}$, Robin Tibor Schirrmeister$^{2}$, Valeriia Cherepanova \footnote{The work does not relate to the author's position at Amazon.} $^{3}$, Chinmay Hegde$^{1}$, \\
\textbf{Frank Hutter$^{2}$, Micah Goldblum$^{1}$, Niv Cohen%
\thanks{Equal advising. Correspondence to: \texttt{bf996@nyu.edu}.
}
$^{1}$, Colin White$^{\dagger 4}$}
    \vspace*{2mm} \\
    $^1$ New York University, $^2$ University of Freiburg, $^3$ University of Maryland, $^4$ Abacus.AI
  }

\begin{document}

\maketitle

\begin{abstract}
While tabular classification has traditionally relied on from-scratch training, a recent breakthrough called prior-data fitted networks (PFNs) challenges this approach. Similar to large language models, PFNs make use of pretraining and in-context learning to achieve strong performance on new tasks in a single forward pass. However, current PFNs have limitations that prohibit their widespread adoption. Notably, TabPFN achieves very strong performance on small tabular datasets but is not designed to make predictions for datasets of size larger than 1000. In this work, we overcome these limitations and substantially improve the performance of PFNs via context optimization. We introduce TuneTables, a parameter-efficient fine-tuning strategy for PFNs that compresses large datasets into a smaller learned context. We conduct extensive experiments on nineteen algorithms over \ndatasets{} datasets and find that TuneTables achieves the best performance on average, outperforming boosted trees such as CatBoost, while optimizing fewer than 5\% of TabPFN's parameters. Furthermore, we show that TuneTables can be used as an interpretability tool and can even be used to mitigate biases by optimizing a fairness objective. We open-source our code and raw results at \url{https://github.com/penfever/TuneTables}.
\end{abstract}
\section{Introduction} \label{sec:introduction}

Tabular data, or data organized into rows and columns consisting of distinct features, are the oldest and one of the most ubiquitous types of data in machine learning in practice \citep{shwartz2022tabular,borisov2021deep}.
Tabular data has numerous applications across medicine \citep{johnson2016mimic,ulmer2020trust}, online advertising \citep{richardson2007predicting,mcmahan2013ad,guo2017deepfm}, finance \citep{arun2016loan,clements2020sequential}, and other areas \citep{chandola2009anomaly,buczak2015survey,urban2021deep}.

Competitive classification algorithms for tabular data include gradient-boosted decision trees \citep{chen2016xgboost,prokhorenkova2018catboost} and deep neural networks \citep{somepalli2021saint,gorishniy2021revisiting,kadra2021well}. Both approaches fit their respective models on a labeled dataset containing samples from a distribution reflecting the task at hand.
A recent breakthrough, prior-data fitted networks (PFNs)  \citep{hollmann2022tabpfn,muller2022transformers}, are are a specific type of neural process which learn to perform approximate Bayesian inference in a single forward pass using in-context learning~\cite{luo2018neural}. PFNs do not require optimizing parameters or fitting a model on downstream training data, instead feeding training data into the context and conditioning on it. 
In particular, TabPFN achieved state-of-the-art classification on small tabular datasets \citep{hollmann2022tabpfn,mcelfresh2023neural}.

The in-context learning approach of PFNs parallels that of large language models (LLMs) \citep{zhao2023survey}. Both approaches can be viewed as approximate Bayesian inference, whether implicitly \citep{xie2022explanation} or explicitly \citep{muller2022transformers}. While researchers have successfully used various context optimization strategies for enhancing LLM performance \citep{liu2023pre}, no prior work has studied context optimization strategies for PFNs. 
Furthermore, although TabPFN achieves very strong performance on small datasets, its limitations currently prohibit its widespread adoption: it only runs on datasets whose number of training samples, number of features, and number of classes are at most 1000, 100, and 10, respectively.

\begin{figure*}
    \centering
    \includegraphics[width=0.99\textwidth]{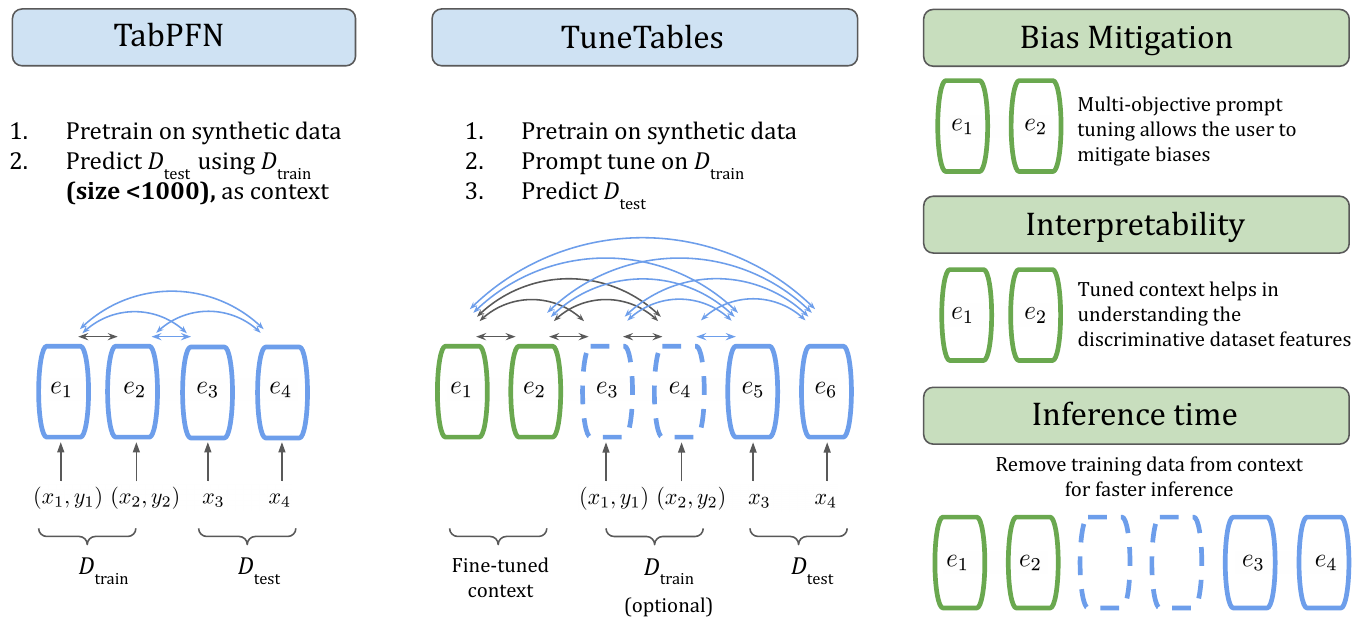}
    \caption{
    \textbf{TuneTables: a novel prompt-tuning technique for prior-data fitted networks. } 
     TuneTables performs prompt tuning on a pre-trained prior-fitted network (TabPFN) to distill real-world datasets into learned embeddings, allowing for stronger performance and faster inference time than TabPFN in many cases. TuneTables also expands the capabilities of pre-trained PFNs; by way of example, we demonstrate its effectiveness for bias mitigation, and as an interpretability tool.
    }
    \label{fig:overview}
\end{figure*}

In this work, we perform the first investigation into context optimization strategies for PFNs, allowing us to substantially improve their performance when scaled to large datasets. Specifically, we introduce \textbf{TuneTables}, a novel parameter-efficient fine-tuning technique for PFNs that compresses large datasets into a smaller learned context (\cref{fig:overview}). 

We conduct an extensive empirical investigation into TuneTables' performance on the TabZilla Benchmark Suite, the largest benchmark considered by recent tabular data literature \citep{mcelfresh2023neural}.

Over \ndatasets{} datasets and 19 algorithms, we find that TuneTables achieves the best average performance and is the best-performing method on 30 of them.

Because TuneTables effectively compresses the contents of large tabular datasets into the tuned prompt, no training data is needed in the context during inference, significantly speeding up inference time (similar to works on neural processes; see \cref{sec:related_work}). We also show that the learned prompt can be used as a tool for interpretability. Finally, we show how to use TuneTables for multi-objective optimization, such as optimizing both accuracy and fairness simultaneously, allowing users to mitigate biased predictions of a pretrained PFNs with just a lightweight tuning procedure.
We open-source our code and raw results at \url{https://github.com/penfever/TuneTables}.

\noindent\textbf{Our contributions.} We describe our main contributions below.
\begin{itemize}[topsep=2pt, itemsep=2pt, parsep=0pt, leftmargin=5mm]
    \item We introduce \textbf{TuneTables}, a parameter-efficient fine-tuning technique for PFNs. TuneTables achieves the highest number of wins (30) when compared to 19 algorithms over 98 datasets, while requiring less inference time than TabPFN.
    \item We show how to use prompt tuning for \textbf{multi-objective optimization}, such as optimizing both accuracy and fairness simultaneously, allowing users to mitigate biased predictions of a pretrained PFNs with just a lightweight tuning procedure.
    \item We show that TuneTables' condensed representations can be used as an \textbf{interpretability} tool. 
    \item We conduct an extensive study on context optimization strategies for PFNs by performing an ablation study on TuneTables, as well as studying sketching and feature selection techniques. 
   \item In order to better manage the limitations of our method, we introduce TuneTables-medium and TuneTables-light, which achieve strong tradeoffs between precision and speed.
\end{itemize}

\begin{comment}

\end{comment}

\newcommand{\xtest}{x_{\text{test}}}%
\newcommand{\ytest}{y_{\text{test}}}%
\newcommand{\Dtrain}{D_{\text{train}}}%
\newcommand{\Dtest}{D_{\text{test}}}%
\newcommand{\Dtune}{D_{\text{tune}}}%
\newcommand{\Dcontext}{D_{\text{context}}}%
\newcommand{\Dcompact}{D_{\text{compact}}}%

%\section{Background and Methodology} 
\section{Background} \label{sec:background}

\paragraph{PFNs: review and limitations.} 
In this section, we give a background on PFNs and discuss their limitations. For a complete description, see \citep{muller2022transformers,hollmann2022tabpfn,nagler2023statistical}.
Assume that we have a classification problem with features $\mathcal{X}\subseteq\mathbb{R}^d$ and labels $\mathcal{Y}$.
Given a dataset $D=\Dtrain\cup\Dtest$, where $\Dtrain=\{(x_1,y_1),\dots,(x_n,y_n)\}$ and $\Dtest=\{(\xtest,\ytest)\}$,
our goal is to predict the conditional class probabilities $p(\cdot\mid\xtest)$.
In the Bayesian framework for supervised learning, the mechanism for generating the data distribution is a hypothesis $\varphi$, drawn from $\Phi$, the space of all hypotheses. $\Phi$ encodes our prior beliefs on the system before observing data. In this framework, datasets $D$ are generated by first drawing $\varphi\sim\Phi$, and then drawing $\emph{i.i.d.}$\ samples according to $\varphi$. 
The posterior predictive distribution (PPD) for a test sample $\xtest$ is the label distribution $p(\cdot\mid\xtest,\Dtrain)$ that follows from our prior.
We can obtain the PPD by integrating over the space of hypotheses $\Phi$:

\begin{equation} \label{eq:ppd}
p(y\mid x,D)\propto \int_\Phi p(y\mid x,\varphi) p(D\mid\varphi)p(\varphi)d\varphi.
\end{equation}

A PFN is a transformer-based architecture trained to approximate the PPD via \emph{synthetic prior-fitting}. Given a prior, we first sample hypotheses $\varphi\sim p(\varphi)$ and then synthetic datasets $D\sim p(D\mid\varphi)$.
We optimize the parameters of the PFN by predicting the class labels of $\Dtest\subseteq D$, conditioned on $\Dtrain=D\setminus\Dtest$.
We compute the loss by:
\begin{equation}
\mathcal{L}_{\text{PFN}}= \E_{D\sim p(D)} \left[ -\log q_\theta(\ytest\mid\xtest,\Dtrain) \right],
\end{equation}
for simplicity assuming all training and test sets are size $n$ and 1, respectively.
We then approximately solve this optimization problem,
%\begin{equation}
%%\hat\theta = \argmax_\theta \E_{D\sim p(D)} \left[ -\log q_\theta(\ytest\mid\xtest,\Dtrain) \right],
$\hat\theta = \argmin_\theta \mathcal{L}_{\text{PFN}}$,
%\end{equation}
allowing $q_\theta$ to approximate \cref{eq:ppd}: 
\begin{equation}
q_\theta(\ytest\mid\xtest,\Dtrain) \approx p(\ytest \mid \xtest,\Dtrain).
\end{equation}

\noindent\textbf{Scaling challenges.}
While PFNs, specifically TabPFN, have shown remarkable success in classification by in-context learning, several important obstacles constrain their more widespread adoption: % as an alternative to GBDTs:

\begin{enumerate}[topsep=2pt, itemsep=2pt, parsep=0pt, leftmargin=5mm]
    \item \textbf{PFNs only accept a fixed number of features.} The current design of PFNs fixes the quantity of features at the time of pretraining. This quantity cannot be changed without retraining the PFN. %For TabPFN, the maximum number of features is 100.
    \item \textbf{PFNs scale poorly with the dataset size}. While PFN accuracy can improve with more real-data samples at inference time \cite{hollmann2022tabpfn,nagler2023statistical}, the memory requirements  scale  with the context length, making extensions beyond
    a certain number of samples impractical.
    \item \textbf{PFNs only select from a fixed number of classes}. The MLP decoder head co-trained with the PFN fixes the number of classes that can be identified at test time.
\end{enumerate}

\paragraph{A motivating example.}  
In \cref{fig:scatterplots} (left), we present a comparison between CatBoost and TabPFNs3000 from  \citep{mcelfresh2023neural}, a version of TabPFN that, for datasets above 3000 data points, uses a random subset of 3000 data points, for fitting; this version also evaluates 30 feature subsets based on mutual information (requiring 30$\times$ more time).
We ablate our feature and sample subselection strategies for TabPFNs3000 in Appendix \cref{tab:ensembling-ablation}.
Although TabPFNs3000 performs very well on datasets with fewer than 1000 datapoints and 100 features, it significantly underperforms CatBoost beyond those constraints.

\paragraph{Approach.}We propose sketching, feature selection, and fine-tuning as an attempt to remedy \emph{(1)} and \emph{(2)}. Then in \cref{sec:tunetables}, we describe novel prompt-tuning and class-extension strategies to create TuneTables, a robust classifier which remedies \emph{(1)}-\emph{(3)}.

\subsection{Classical sketching and feature selection}
\label{sec:sketching}

\paragraph{Sketching.}
The number of samples that a PFN can handle is limited to around 3000 by conventional GPU sizes.
However, in the real world, datasets are often much larger. Going forward, we refer the maximum allowable context size of the PFN as $n$, and to the size of the real-world dataset as $N$.

Given a training dataset $\Dtrain$ of size $N >> n$, one option is to select a representative subset of the dataset, $\Dcompact\subseteq\Dtrain$, to use as the context. 
In general, researchers have studied a variety of data summarization techniques, often called \emph{sketching}, for tabular data \citep{munteanu2018coresets}.
In the context of a pretrained PFN $q_\theta$, the goal is to find a sketching function $s:\R^{N\times d}\mapsto\R^{n\times d}$ such that
\begin{align}
&\E_{D\sim p(D)}\left[-\log q_\theta(\ytest\mid\xtest,s(\Dtrain))\right] \label{eq:sketching} \\
\approx~
&\E_{D\sim p(D)}\left[-\log q_\theta(\ytest\mid\xtest,\Dtrain)\right]. \nonumber
\end{align}
Here, we consider three sketching methods for TabPFN: \emph{random}, in which we select a random subset of $n$ datapoints from the full training set; $k$-\emph{means}, in which we compute the $n$-means clustering \citep{ahmed2020k} of $\Dtrain$ and select the $n$ centers; and \emph{CoreSet}, in which we compute a core-set of size $n$ \citep{agarwal2005geometric}.

\paragraph{Feature selection.}
In addition to the limitation on dataset size, PFNs, such as TabPFN, also impose a limitation on the number of features $d$.
Similar to sketching, given a dataset with $D >> d$ features, we can perform feature selection or summarization in order to adhere to the constraint (formally, finding a function that reduces the feature dimension and approximates an expression similar to \cref{eq:sketching}).
Feature selection is a critical part of tabular classification, and there are several popular feature selection methods \citep{chandrashekar2014survey,cherepanova2023performance}.
We investigate three different methods: \emph{random}, in which we randomly select a set of $d$ features; \emph{mutual information}, in which we select $d$ features with the highest mutual information of the target dataset \citep{vergara2014review}; and \emph{principal component analysis (PCA)}, in which we take the $d$ first principal components.
In \cref{sec:experiments}, we find that all sketching and feature selection methods plateau in performance well before approaching parity with GBDTs, which motivates our investigation of new scaling techniques.

\subsection{Fine-tuning}
\label{sec:ft-pfn}
We conclude this section with a discussion of another potential approach for scaling PFNs: \textbf{fine-tuning}.
In this approach, given a dataset of size $N>>n$, we use gradient descent to continue training \emph{all parameters} of the PFN.
However, we show in \cref{sec:experiments} that fine-tuning takes up considerable memory resources while still not achieving competitive accuracy on large datasets; we postulate that for most datasets, the synthetic prior of TabPFN is more robust to overfitting than the actual training data; fine-tuning overfits to the validation set. This is borne out by the observation that when fine-tuning TabPFN, validation accuracy continues to improve even as test accuracy declines. To remedy this, we present new parameter-efficient solutions for scaling PFNs in the next section.

\section{TuneTables}  \label{sec:tunetables}

Motivated by the strong performance of prompt tuning, we introduce \textbf{TuneTables}, a new tabular classification algorithm. Using a pretrained PFN (TabPFN, in our experiments) as a base model, TuneTables overcomes the limitations of PFNs, allowing them to be applied to any tabular classification problem.

For the full details of the algorithm, please refer to \cref{app:pt-experiments}. We summarize here: \emph{(a)} if a dataset is small enough to run with the original zero-shot version of TabPFN, we include it in our search, as the TabPFN is already highly optimized for such datasets; The reason why we do not only use TabPFN is that the exact size at which TuneTables outperforms TabPFN is dataset-dependent (with an average transition of around 800 samples). \emph{(b)} if there are more than 100 features, we perform grid search over a set of feature subselection methods and select the one which performs best on the validation set;
\emph{(c)} orthogonally, if there are too many labels, we fit a new decoder to a frozen TabPFN;
\emph{(d)} we optimize over a search space of tuned prompts, both with and without real-data context during training, fitted to the dataset;
\emph{(e)} we report the best-performing model according to accuracy.

\begin{figure*}[t]
    \centering
    \includegraphics[width=0.85\textwidth]{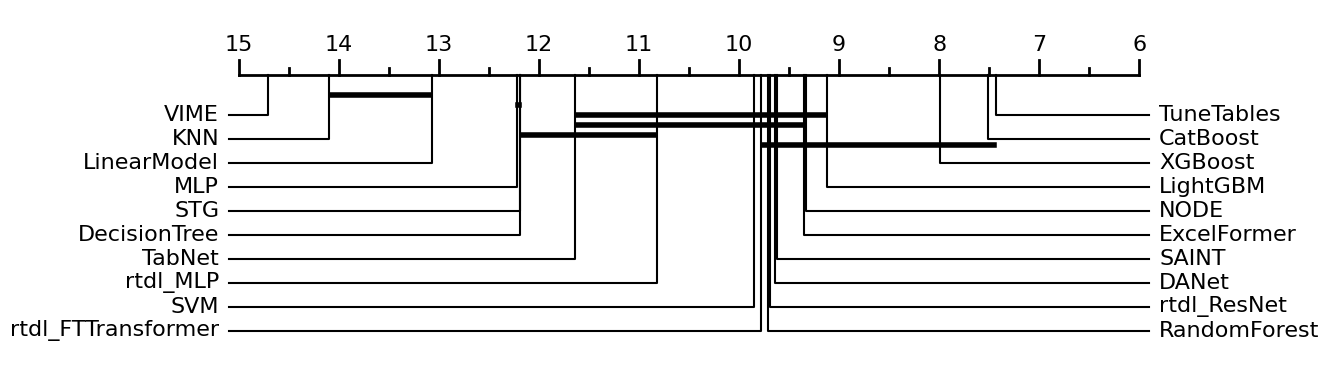}
    \caption{\textbf{TuneTables and state-of-the-art tabular models.} A critical difference plot according to mean accuracy rank across the \ndatasets{} datasets in Table 1 of \cite{mcelfresh2023neural}. Algorithms which are \emph{not significantly different} ($p>0.05$) are connected with a horizontal black bar. TuneTables achieves the highest mean rank of any algorithm. 
    }
    \label{fig:critical_difference}
\end{figure*}

\paragraph{Prompt tuning as a scalable context for PFNs.}Motivated by the limitations of sketching for large contexts, we explore \emph{soft prompt tuning} as an alternative. 
In soft prompt tuning \citep{lester2021power}, given a dataset $D=\Dtrain\cup\Dtest$, a parameterized matrix $\Dtune^{p \times e}$ is prepended to the input embedding $\Dtrain^{n \times e}$, where $e$ is the transformer embedding dimension and $p$ is a hyperparameter--the size of the tuned prompt. 

The paper \citep{lester2021power} demonstrates the effectiveness of soft prompt tuning for NLP tasks by prepending a small task-specific prompt ($p \approx 5$). These task-specific learned tokens prove effective at the extremely large model scales commonly encountered in NLP. Somewhat surprisingly, 
we show that prompt tuning is effective at similar scales of $p$ even for the much smaller tabular data models (see \cref{app:pt-experiments} and \cref{app:interpretability}). However, prompt tuning increases in effectiveness when $p$ is larger.

\paragraph{Soft prompt tuning for tabular data.}
Unlike in NLP, transformers for tabular data, including PFNs, generally accept two input embeddings; a continuous-valued embedding $\Dtrain{_X}$, and a categorically-valued $\Dtrain{_y}$ which is passed through directly. We adapt the method of \cite{lester2021power} by fitting the parameters of 
$D_{\text{tune}\,X}^{p \times e}$ to $\Dtrain{_X}$ and randomly initializing 
$D_{\text{tune}\,y}^{p \times 1}$ with an equal number of labels from each class in $\Dtrain$. These synthetic datapoints are optimized on the entire labeled set, and therefore give PFNs access to a much larger training set not accessible by existing methods.

We further adjust the method of \cite{lester2021power} to allow for the possibility that $\Dtune$ has learned, in essence, a distilled version of $\Dtrain$; at test time, we evaluate two settings, hereafter referred to as $C$ (`context') and $NC$ (`no context'). In the $C$ setting, following \cite{lester2021power}, we have $\Dtune^{p \times e}$ prepended to the input embedding $\Dtrain^{n \times e}$. In the $NC$ setting, we provide only $\Dtune^{p \times e}$ at test time. In \cref{sec:experiments}, we empirically evaluate both approaches. We ablate specifics of our implementation choices in \cref{app:pt-experiments}. Unlike prompt tuning for NLP, we show that the $NC$ setting is often competitive with, if not better than, the $C$ setting. We also ablate this choice during training; see \cref{app:pt-experiments} for implementation specifics.

\paragraph{Extending the number of predicted classes.}TabPFN uses a pretrained transformer, with a two-layer MLP as a final classifier. The pretraining procedures limit the na\"ive use of TabPFN to classification tasks with at most 10 classes. Yet, in many cases, datasets of interest might contain a larger number of classes, which would require pretraining a new PFN from scratch.

Following the method of last-layer retraining \cite{kirichenko2022last,le2023last}, for such datasets, we fit a PFN to new posteriors given by real-world classification datasets with more than $10$ classes, freezing all weights except for those of the decoder MLP and the tuned prompt (see \cref{app:pt-experiments} for more implementation details). Even after this modification, our method remains highly parameter-efficient, optimizing fewer than 5\% of TabPFN's parameters.
\section{Experiments} \label{sec:experiments}

\begin{figure*}[t]
    \centering
    \includegraphics[width=0.32\textwidth]{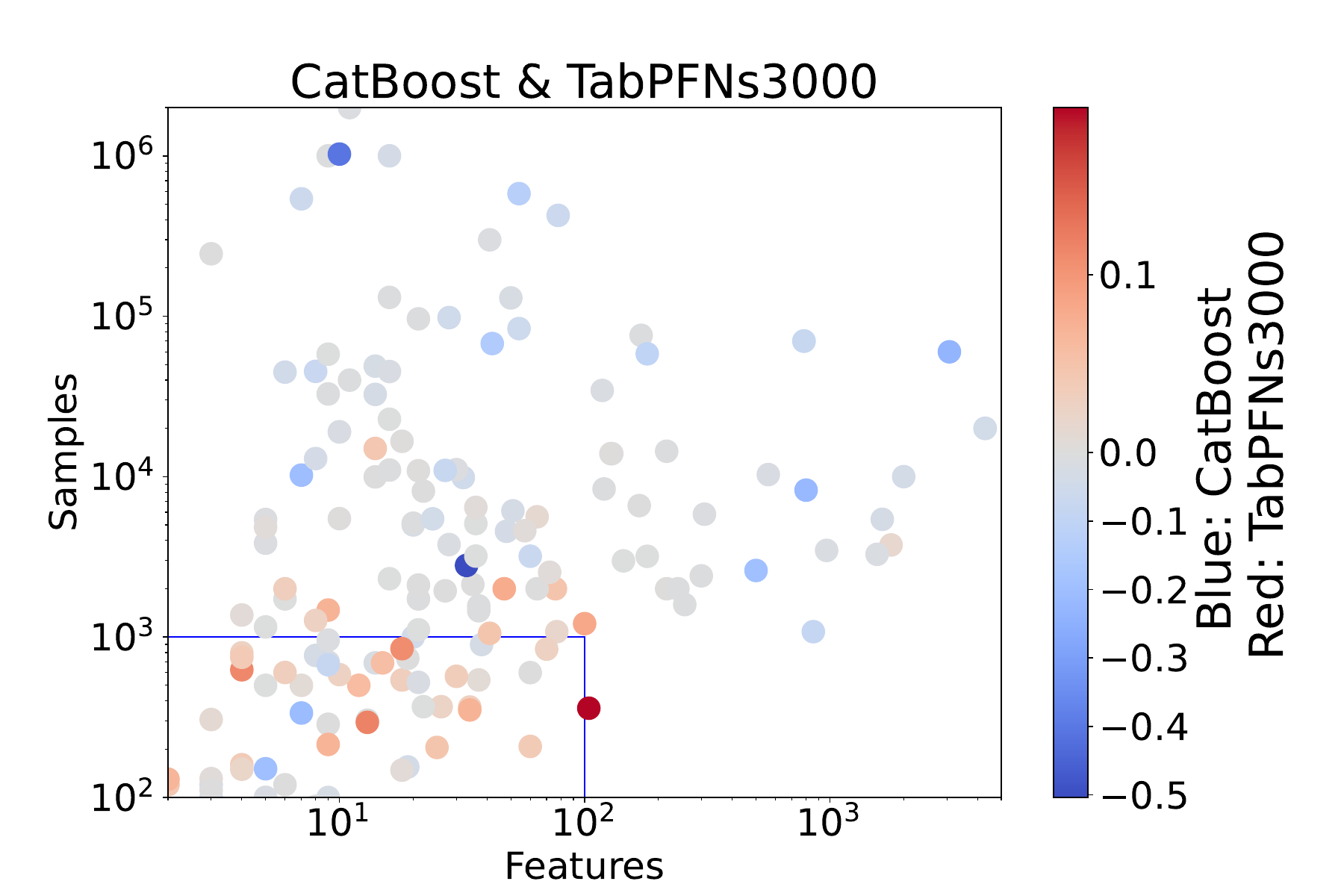}
    \includegraphics[width=0.32\textwidth]{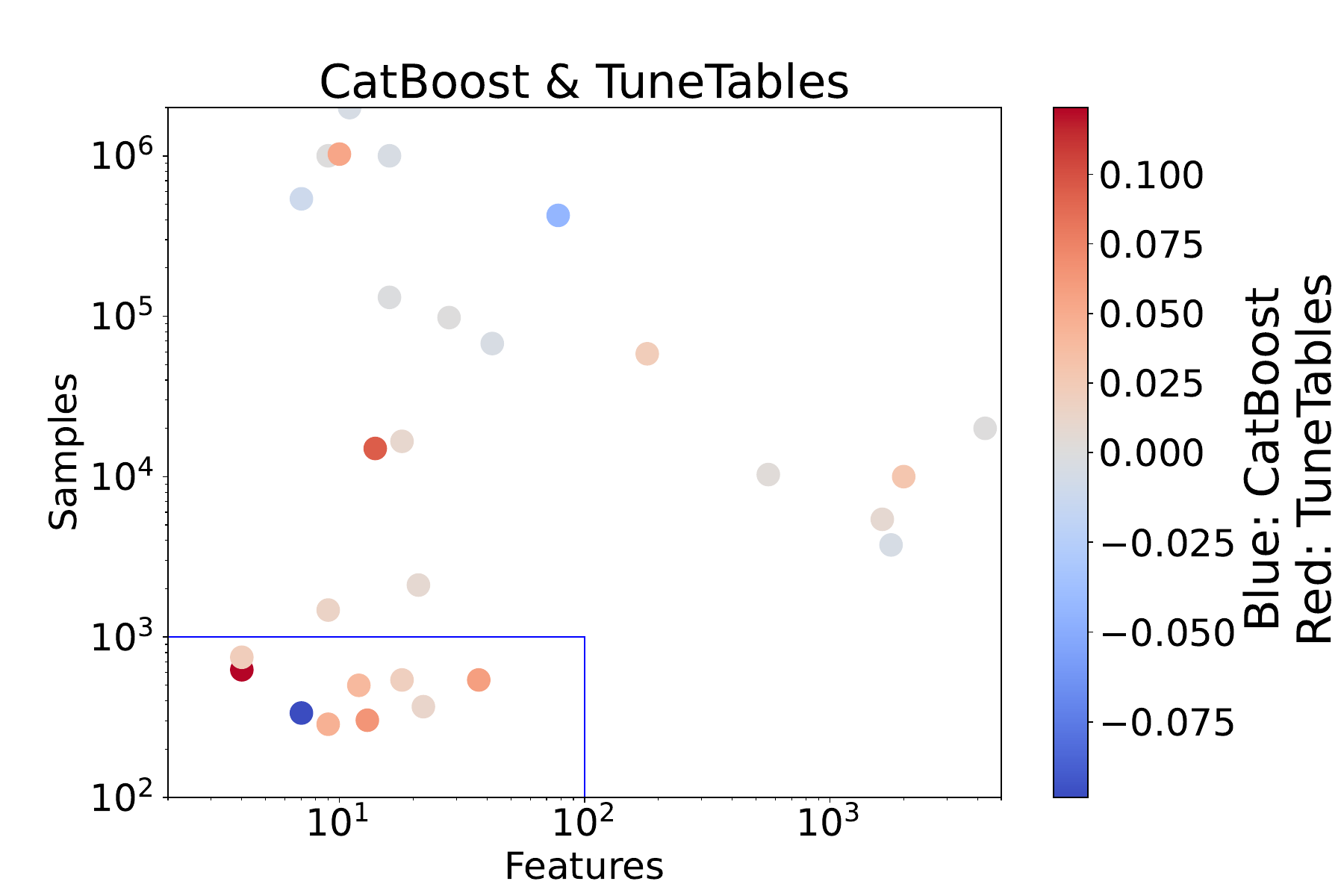}
    \includegraphics[width=0.32\textwidth]{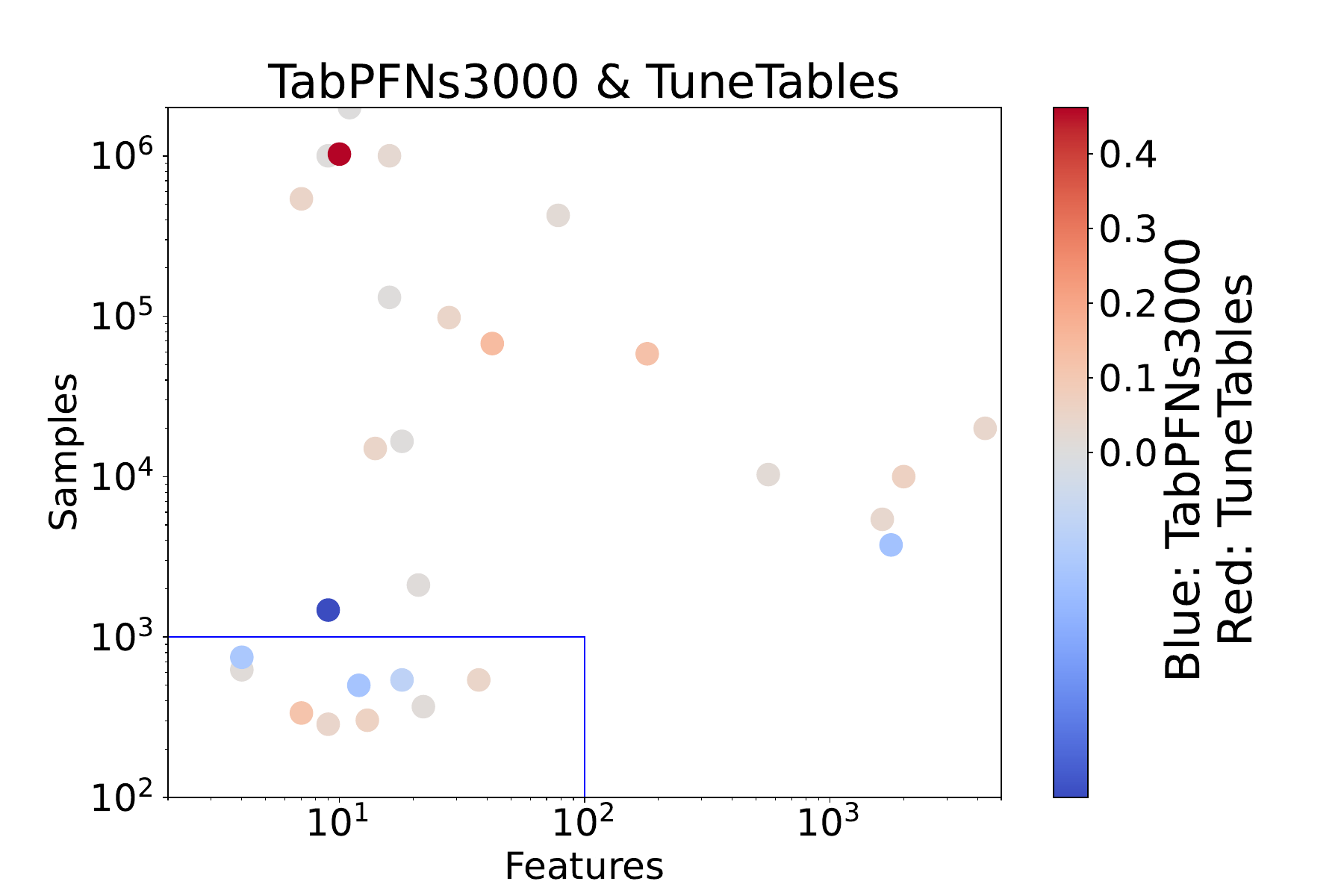}
    \caption{\textbf{TuneTables addresses TabPFN's limitations.} 
    (Left) Motivating example (using the subset of  \citep{mcelfresh2023neural}on which both CatBoost and TabPFNs3000 report results): TabPFNs3000 is best on small datasets, but when scaled past 3000 datapoints and 100 features, TabPFNs3000 significantly underperforms.
    (Middle) CatBoost vs.\ TuneTables on \newbenchmark: By contrast, TuneTables is competitive with CatBoost on all datasets, mitigating the limitations of TabPFN.
    (Right) TabPFNs3000 vs.\ TuneTables on \newbenchmark: TuneTables  outperforms TabPFNs3000 on datasets with a high number of datapoints or features. The colorbar on the y axis represents the comparative change in per-dataset accuracy between two algorithms (A: blue, B: red). Positive numbers represent the absolute gain in accuracy of B w.r.t. A, negative numbers represent the absolute gain in accuracy of A w.r.t. B.
    }
    \label{fig:scatterplots}
\end{figure*}

\paragraph{Algorithms and datasets used.}
We compare TuneTables to nineteen algorithms, including three GBDTs: 
CatBoost \citep{prokhorenkova2018catboost}, 
LightGBM \citep{ke2017lightgbm}, and
XGBoost \citep{chen2016xgboost};
11 neural networks:
DANet \citep{chen2022danets},
FT-Transformer \citep{gorishniy2021revisiting},
two MLPs \citep{gorishniy2021revisiting},
NODE \citep{Popov2020Neural},
ResNet \citep{gorishniy2021revisiting},
SAINT \citep{somepalli2021saint},
STG \citep{yamada2020feature},
TabNet \citep{arik2021tabnet},
TabPFN \citep{hollmann2022tabpfn},
VIME \citep{yoon2020vime}, and
ExcelFormer \citep{chen2024excelformerneuralnetworksurpassing};
and five baselines:
Decision Tree \citep{quinlan1986induction},
KNN \citep{cover1967nearest},
Logistic Regression \citep{cox1958regression},
Random Forest \citep{liaw2002classification}, and
SVM \citep{cortes1995support}.
For all algorithms, we use their official implementation; see \cref{app:experiments} for more details.
We also compare to TabPFNs3000.
We run the algorithms on the TabZilla Benchmark Suite introduced in \cite{mcelfresh2023neural}. This suite consists of \ndatasets{} classification datasets from OpenML \citep{vanschoren2014openml} with a diversity of sizes and number of features \citep{mcelfresh2023neural}.
See \cref{tab:datasets} in \cref{app:experiments} for a list of all datasets with their statistics.

\paragraph{Experimental setup.}
For all algorithms other than TuneTables, we perform light hyperparameter tuning by running one default setting and 29 iterations of random search using Optuna \citep{akiba2019optuna}); see \cref{app:experiments} for details. Following \citep{mcelfresh2023neural}, all of the algorithms come with their default set of hyperparameters used in the official implementation, and we used all of these settings. 
For TuneTables, we optimize via a grid search described in \cref{app:pt-experiments}.

We fit for up to 100 epochs  with early stopping. For all algorithms, we report the test performance of the hyperparameter set with the best performance on the validation set and cross-validate on three train/test folds from OpenML. We conduct our experiments on an NVIDIA L4 TPU with 24GB VRAM.
We summarize our results across datasets by reporting mean accuracy as well as the mean normalized accuracy (after Z-score normalization) for each algorithm.

\paragraph{Algorithm comparison.}
We compute \emph{statistically significant} performance differences among algorithms averaged across all datasets, as done in prior work \citep{kadra2021well}.
We first use a Friedman test to check whether performance differences between all algorithms are significant \citep{friedman1937use}. We reject the null hypothesis for $p<0.05$.
Then we use a Wilcoxon signed-rank test to check which pairs of algorithms have significant performance differences \citep{conover1999practical}. 
We use a Holm-Bonferroni correction to account for multiple comparisons \citep{holm1979simple}. 
See \cref{fig:critical_difference}.
We find that TuneTables achieves the highest average rank, although the difference among the top three algorithms is not statistically significant. In \cref{tab:less_than_50k}, we present the accuracy, rank, and Z-score of all algorithms, averaged across all datasets, and find that TuneTables performs very well on all metrics but runtime.

\paragraph{Large datasets.} One limitation of the TabZilla Benchmark Suite is that even the largest dataset in the comparison contains only 45,211 samples. This scale is modest by the standards of modern tabular problems. In order to better understand the performance of TuneTables on extremely large datasets, we curate from \cite{mcelfresh2023neural} a novel benchmark, \textit{\newbenchmark}, consisting of 29 datasets with up to 1\,900\,000 samples, and 7200 features. We curate the  datasets from OpenML, omitting image classification datasets. Since TabPFN generally outperforms boosted trees on these smaller datasets, and TuneTables extends TabPFN, we also heuristically select smaller datasets to \newbenchmark so as not to favor either TabPFN or boosted trees. TuneTables achieves the highest average accuracy of any algorithm, and achieves the best performance on \texttt{poker-hand}, a dataset of size 1\,025\,009. The complete results can be found in Appendix \cref{tab:full_results} and  \cref{fig:scatterplots} (right). In order to assess the performance of TuneTables on datasets with many classes, we curate another subset of \cite{mcelfresh2023neural}, presenting results on fifteen datasets with up to 100 classes. 
In appendix \cref{tab:large_cardinality} we show that despite the large divergence from the PFN’s pretraining, TuneTables achieves a mean rank of 2.52, ahead of CatBoost and a ResNet, second only to XGBoost, whose mean rank is 2.0.

\begin{table*}[t]
\caption{
\textbf{TuneTables matches SOTA algorithms on 98 datasets.}
In this table, we compare algorithms over the \ndatasets{} datasets in the TabZilla benchmark suite from \cite{mcelfresh2023neural}. For each algorithm, we compute its mean accuracy and mean rank in terms of accuracy.
We also compute the mean Z-score, computed by normalizing the set of results on each dataset (by mean 0 std.\ 1), so that each dataset has the same weight, and averaging each algorithm's normalized performances. Std.\ Z-Score is computed with respect to random splits and averaged across datasets. Num.\ wins values are averaged over three splits per dataset.
}
\label{tab:less_than_50k}
\centering
\resizebox{.97\textwidth}{!}{%
\begin{tabular}{lrrrrrr}
\toprule
        \textbf{Model} & \textbf{Mean Acc.} & \textbf{Mean Rank} & \textbf{Mean Z-Scores} & \textbf{Std Z-Scores} & \textbf{Med Z-Scores} & \textbf{Num.\ Wins} \\ \hline
    TuneTables & \textbf{0.860} & \textbf{7.435} & 0.494 & \textbf{0.624} & 0.490 & \textbf{30} \\
    CatBoost & 0.856 & 7.514 & \textbf{0.496} & 0.669 & \textbf{0.566} & 13 \\
    XGBoost & 0.854 & 7.991 & 0.411 & 0.783 & 0.533 & 16 \\
    ExcelFormer & 0.847 & 9.349 & 0.212 & 0.863 & 0.384 & 7 \\
    LightGBM & 0.845 & 9.122 & 0.284 & 0.894 & 0.431 & 20 \\
    RandomForest & 0.843 & 9.713 & 0.206 & 0.776 & 0.374 & 4 \\
    SAINT & 0.840 & 9.619 & 0.132 & 0.985 & 0.337 & 10 \\
    DANet & 0.840 & 9.646 & 0.209 & 0.763 & 0.345 & 3 \\
    rtdl\_ResNet & 0.839 & 9.691 & 0.175 & 0.798 & 0.356 & 5 \\
    rtdl\_FTTransformer & 0.838 & 9.782 & 0.198 & 0.786 & 0.273 & 8 \\
    NODE & 0.838 & 9.335 & 0.224 & 0.695 & 0.357 & 3 \\
    SVM & 0.835 & 9.847 & 0.121 & 0.812 & 0.318 & 12 \\
    DecisionTree & 0.823 & 12.192 & -0.308 & 1.215 & -0.038 & 4 \\
    rtdl\_MLP & 0.813 & 10.825 & -0.019 & 0.907 & 0.223 & 1 \\
    STG & 0.810 & 12.196 & -0.290 & 1.051 & -0.057 & 5 \\
    TabNet & 0.804 & 11.643 & -0.216 & 1.104 & 0.075 & 5 \\
    MLP & 0.802 & 12.220 & -0.232 & 0.845 & -0.105 & 1 \\
    LinearModel & 0.793 & 13.069 & -0.520 & 1.233 & -0.345 & 5 \\
    KNN & 0.781 & 14.101 & -0.727 & 1.187 & -0.608 & 0 \\
    VIME & 0.756 & 14.711 & -0.849 & 1.192 & -0.694 & 5 \\
\bottomrule
\end{tabular}
}
\end{table*}

\paragraph{Runtime comparison.} We divide our consideration of runtime into inference and training. At inference time, TuneTables is around 9x faster than TabPFNs3000, on average; see Appendix \cref{tab:inference} for the per-dataset details.
However, the end-to-end runtime of TuneTables is over 7x that of CatBoost and XGBoost, and also slower than TabPFNs3000, because of the increased \emph{training time}. 

In order to better understand the trade-off between accuracy and runtime, we introduce efficient variants of our method. 
\emph{TuneTables-medium} utilizes a more efficient adaptive sequence size (i.e., the number of real data points received at train time) which scales with the size of the dataset, validates on a subset of the available validation set when the validation set is large, employs lower patience for early stopping, omits a zero-shot TabPFN grid search over 30 random seeds which TuneTables-standard uses to find more optimal feature selection subsets, and, most impactfully, omits ensembling for datasets larger than a cutoff hyperparameter (which we fix at 150\,000 samples). \emph{TuneTables-light} includes all of the efficiency modifications of TuneTables-medium, trains for just one epoch, and uses TabPFN zero-shot to preselect the feature selection method rather than performing a grid search using TuneTables itself.
In \cref{tab:runtimes_summary}, we compare these lighter methods to TuneTables-standard on the \newbenchmark benchmark. TuneTables-medium decreases runtime by 72\% compared to TuneTables-standard, while the accuracy decreases by less than a quarter of a percent.
Furthermore, the runtime of TuneTables-light is comparable to CatBoost; however, performance also degrades by 5\% when going from TuneTables-medium to TuneTables-light. Still, TuneTables-light shows stronger performance than TabPFNs3000 (78.7\% vs. 78.1\% accuracy) while having a lower inference time.

\begin{table}[t]
    \caption{\textbf{TuneTables-medium and TuneTables-light are substantially faster with only a modest decrease in accuracy.}
    We compare the average accuracy and runtime (in seconds) of three versions of TuneTables and find that the medium and light versions of the algorithm are substantially faster on large datasets; we also find that TuneTables-medium sacrifices little accuracy.
    \label{tab:runtimes_summary}
    }
        \centering
    \resizebox{.8\textwidth}{!}{%    
    \centering
    \begin{tabular}{lcccccc}
        \toprule
        & \textbf{TuneTables} & \textbf{TuneTables-medium} & \textbf{TuneTables-light} \\
        \midrule
        Avg.\ acc (\newbenchmark, size $<50$K) & 0.831 & 0.830 & 0.810 \\
        Avg.\ runtime (\newbenchmark, size $<50$K) & 1325 & 1026 & 450 \\
        \midrule
        Avg.\ acc (\newbenchmark, all datasets) & 0.830 & 0.828 & 0.787 \\
        Avg.\ runtime (\newbenchmark, all datasets) & 6975 & 1908 & 486 \\
        \midrule
        Avg.\ acc (TabZilla, all datasets) & 0.861 & 0.855 & 0.854 \\
        Avg.\ runtime (TabZilla, all datasets) & 573 & 305 & 196 \\
        \bottomrule
    \end{tabular}
    }
\end{table}

\paragraph{Ablations.}
In order to better understand the significance of the changes we introduce in our method, we separately ablate the tuned prompt size and the use of ensembles in \cref{tab:ablation} and \cref{tab:ensembling-ablation}, finding that smaller datasets pair well with smaller prompts and rarely benefit from ensembling.

We also compare TuneTables with and without keeping the real data as additional context (referred to as C and NC, respectively, in \cref{sec:tunetables}); see Appendix \cref{tab:ablation}.
We find that smaller datasets cannot always be fully learned by the tuned prompt, but for larger datasets, the tuned prompt alone suffices, and in some cases even outperforms  the real data.

\paragraph{Sketching and feature selection}
Finally, in Appendix \cref{tab:sketching_fs} we give a study on the three sketching and three feature selection techniques described in \cref{sec:background}.
As described earlier, TabPFNs3000's performance when relying on feature selection plateaus well before approaching parity with CatBoost, a top-performing GBDT, on seven very large datasets.

\section{TuneTables Extensions} \label{sec:fairness}

\begin{table*}[t]
\caption{\textbf{TuneTables significantly improves accuracy and demographic parity.} In these multi-objective optimization experiments, we consider prompt tuning for mitigating predictive bias, comparing TabPFN to TuneTables, tuning for accuracy alone vs.\ accuracy and demographic parity.
TuneTables improves over TabPFN with respect to both objectives. 
}
\label{tab:fairness}
\centering
    \resizebox{.75\textwidth}{!}{%    
\begin{tabular}{lcccccccccc}
\toprule
& \multicolumn{2}{c}{Adult} & \multicolumn{2}{c}{Speeddating} & \multicolumn{2}{c}{Compas} & \multicolumn{2}{c}{NLSY} & \multicolumn{2}{c}{Average} \\
\cmidrule(lr){2-3} \cmidrule(lr){4-5} \cmidrule(lr){6-7} \cmidrule(lr){8-9} \cmidrule(lr){10-11}
 & Acc $\uparrow$ & DP $\downarrow$ & Acc $\uparrow$ & DP $\downarrow$ & Acc $\uparrow$ & DP $\downarrow$ & Acc $\uparrow$ & DP $\downarrow$ & Acc $\uparrow$ & DP $\downarrow$ \\
\midrule
TabPFN  & 0.832  & 0.174  & 0.86  & 0.012 & 0.688  & 0.22 & \textbf{0.986} & 0.326 & 0.842 & 0.183 \\
\midrule
TuneTables (Acc)  & \textbf{0.845}  & 0.13 & \textbf{0.865}  & 0.006  & 0.688  & 0.209  & 0.974 & 0.302 & \textbf{0.843} & 0.162 \\
TuneTables (Acc + DP)  & 0.837  & \textbf{0.034}  & 0.863  & \textbf{0.003}  & \textbf{0.693}  & \textbf{0.121} & 0.965 & \textbf{0.277} & 0.840 & \textbf{0.109} \\
\bottomrule
\end{tabular}
}
\end{table*}

\noindent\textbf{Mitigating bias with prompt tuning.} Many real-world applications of machine learning involve a set of protected attributes (such as race or gender) that partition the dataset into groups, in which some  have higher model performance than others.
Removing the sensitive attributes does not fix the algorithmic bias, because the sensitive attributes are often non-trivially correlated with other attributes in the dataset.
Due to this issue, researchers have put in significant effort into mitigating the bias of ML models, with the majority of techniques devising new training strategies \citep{barocas2023fairness,mehrabi2021survey}.

\begin{wrapfigure}{r}{0.5\textwidth} 
    \centering

        \vspace{-10pt} 
\includegraphics[width=0.4\textwidth]{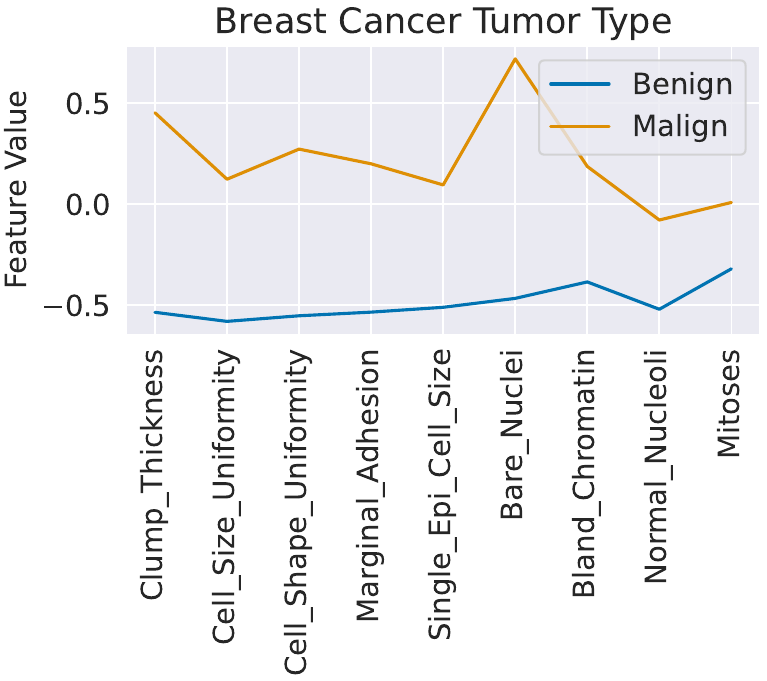}
    \vspace{-5pt} 
    \caption{\textbf{Dataset with high accuracies from just two datapoints}. Shown is a two-example prompt dataset for the breast cancer dataset \citep{misc_breast_cancer_wisconsin_(diagnostic)_17}. Malign class example has higher values for all features than benign class.}
    \label{fig:breast_cancer}
\end{wrapfigure}

Given TabPFN's pretrained nature and considerable retraining cost, the only options for mitigating biased predictions are to run a post-processing routine on the output predictions, which generally do not perform as well as in-processing strategies \citep{savani2020intra}. 
We show how to use prompt tuning to substantially reduce the bias of predictions while also improving accuracy.

We conduct experiments on four datasets widely used for research in fairness: the Adult Census Income database (with sex as the sensitive attribute) \citep{asuncion2007uci}, speed dating (with same race as the sensitive attribute) \citep{zheng2018fairness}, COMPAS (with sex as the sensitive attribute) \citep{angwin2022machine}, and National Longitudinal Survey (with gender as the sensitive attribute) \citep{BLS2023}.
To quantify bias, we use \emph{demographic parity} \citep{verma2018fairness,dwork2012fairness}, which measures the difference in probability of a positive outcome among the protected and unprotected groups. Formally, given protected group $G_1$, unprotected group $G_0$, and protected attribute $x_{\cdot,a}$, it can be computed as
\begin{equation*}
P_{(x_i,y_i)\in G_0}(y_i=1\mid x_{i,a})-P_{(x_i,y_i)\in G_1}(y_i=1\mid x_{i,a}).
\end{equation*}

During prompt tuning, we employ a demographic parity regularizer that aims to minimize the difference in positive outcome probabilities between the two groups:

\begin{equation*}
%L_{DP} =
\left| \sum_{(x_i,y_i) \in G_0} P(y_i=1\mid x_{i,a}) - \sum_{(x_i,y_i) \in G_1} P(y_i=1\mid x_{i,a}) \right|
\end{equation*}

We use the same experimental setup as in \cref{sec:experiments}, except that we report one shift rather than the average of three, and TuneTables is fitted to a single prompt rather than ensembled. We compare the default TabPFN to TuneTables, fine-tuning for accuracy alone vs.\ accuracy and demographic parity. The latter significantly improves demographic parity, compared to the default TabPFN and TuneTables fine-tuned for accuracy, and enhances accuracy relative to the default TabPFN across all datasets but one; see \cref{tab:fairness}.

\noindent\textbf{Summarizing and understanding datasets with prompt tuning.} While we have demonstrated that TuneTables scales and improves the performance of PFNs, now we show that it can also help understand the discriminative features in a dataset. Often, besides a good predictive model for a dataset, users want to gain further insights into the dataset. In prompt tuning, the tuned smaller dataset can be seen as a summary of the complete dataset that emphasizes discriminative features for the given task. As an example, in \cref{fig:breast_cancer} we show that on the Breast Cancer dataset \citep{misc_breast_cancer_wisconsin_(diagnostic)_17}, a prompt with just two synthetic examples is enough to reach high accuracies and at the same time allows an understanding of the predictive features. For example, in the Breast Cancer dataset, the malign example has higher values for all features compared to the benign example, suggesting high feature values indicate malignancy. We give further examples in \cref{app:interpretability}.
\section{Related work} \label{sec:related_work}

% short version:

\paragraph{Neural Processes and Prior-Data Fitted Networks.}
Prior-data Fitted Networks (PFNs) \citep{muller2022transformers,hollmann2022tabpfn} are a recently-proposed paradigm for machine learning, which show that fast approximate Bayesian inference is possible by training a neural network to mimic the posterior predictive distribution (PPD) in a single forward pass using in-context learning \citep{muller2022transformers,muller2023pfns4bo,nagler2023statistical}.

PFNs were shown to yield state-of-the-art empirical performance on small tabular datasets \citep{hollmann2022tabpfn, mcelfresh2023neural}.
PFNs have been used in other applications, including Bayesian optimization \citep{muller2023pfns4bo} learning curve extrapolation \citep{adriaensen2022efficient}, and as foundation models for hypernetworks \citep{müller2023mothernet}.

PFNs are neural processes (NPs) \citep{muller2022transformers}. Recent advances in NP are similar in nature to our work, especially recent works that aim to scale attention-based methods to larger contexts.
Feng et al.\ \citep{feng2022latent} propose Latent Bottlenecked Attentive Neural Processes (LBANPs), a transformer-based neural process which overcomes the quadratic complexity of transformers by encoding the context dataset into a constant number of latent vectors.
Guo et al.\ \citep{guo2022versatile} propose Versatile Neural Processes (VNP), which increases the capability of NPs to handle compex signals by using a new bottleneck encoder.
\citep{rastogi2022semi} introduces semi-parametric inducing point networks (SPIN), which can attend to a training set at inference time with linear complexity via inducing point methods.
\citep{feng2023memory} introduce Constant Memory Attention Block (CMAB), an attention block that is permutation-invariant and has constant memory complexity when computing the output, as well as  Constant Memory Attentive Neural Processes (CMANPs), a NP that requires constant memory.
For additional related work, see \cref{app:related_work}.

\section{Conclusions, limitations, and future work} \label{sec:conclusion}
In this work, we gave the first investigation into context optimization techniques for PFNs, allowing us to substantially improve their performance when scaled to large datasets.
In particular, we introduced TuneTables, which uses a novel prompt-tuning technique to achieve strong performance on large datasets.
We demonstrate that TuneTables mitigates the constraints of TabPFN on the dataset size, number of features, and number of class labels.
Additionally, we use prompt tuning to mitigate bias without retraining TabPFN and as an interpretability tool.
We open-source our code, results, and all other materials needed to reproduce our work.
As PFN models improve, the context optimization techniques explored in our work will allow researchers to further optimize and scale. For example, a next-generation TabPFN might have a longer total context length, and prompt tuning will allow us to push the dataset size even further.

\paragraph{Limitations.} No current TuneTables is on par with GBDTs in terms of both accuracy and runtime simultaneously. We therefore emphasize that while we achieve strong performance metrics, we do not claim practical superiority of our method over gradient boosting, when taking into account training time. However, given the novelty of our method, we expect future research to further improve the accuracy-runtime tradeoff of TuneTables. It also does not improve on TabPFN for small datasets (fewer than 1000 samples, 100 features and 10 classes); we postulate that this is a result of overfitting. % and leave the mitigation of this to future work. 

\paragraph{Future work.} Parameter-efficient fine-tuning can be used with PFNs to ensure high-quality, differentially private predictions, given that the pretraining is done on purely synthetic data, and prompt tuning only updates a small number of parameters.
Low-rank adaptation (LoRA) \citep{hu2021lora} and quantized low-rank adaptation (QLoRA) \citep{dettmers2023qlora} have been used successfully for large language models and would be a promising technique for parameter-efficient fine-tuning of PFNs.
Designing a sparse mixture-of-experts PFN using router networks is another promising technique, due to its success in the field of large language models \citep{jiang2024mixtral}.

\bibliography{main}

\begin{thebibliography}{10}

\bibitem{adriaensen2022efficient}
Steven Adriaensen, Herilalaina Rakotoarison, Samuel M{\"u}ller, and Frank Hutter.
\newblock Efficient bayesian learning curve extrapolation using prior-data fitted networks.
\newblock In {\em Proceedings of the Annual Conference on Neural Information Processing Systems (NeurIPS)}, 2023.

\bibitem{agarwal2005geometric}
Pankaj~K Agarwal, Sariel Har-Peled, Kasturi~R Varadarajan, et~al.
\newblock Geometric approximation via coresets.
\newblock {\em Combinatorial and computational geometry}, 2005.

\bibitem{ahmed2020k}
Mohiuddin Ahmed, Raihan Seraj, and Syed Mohammed~Shamsul Islam.
\newblock The k-means algorithm: A comprehensive survey and performance evaluation.
\newblock {\em Electronics}, 2020.

\bibitem{akiba2019optuna}
Takuya Akiba, Shotaro Sano, Toshihiko Yanase, Takeru Ohta, and Masanori Koyama.
\newblock Optuna: A next-generation hyperparameter optimization framework.
\newblock In {\em Proceedings of the 25th ACM SIGKDD international conference on knowledge discovery \& data mining}, pages 2623--2631, 2019.

\bibitem{angwin2022machine}
Julia Angwin, Jeff Larson, Surya Mattu, and Lauren Kirchner.
\newblock Machine bias.
\newblock In {\em Ethics of data and analytics}, pages 254--264. Auerbach Publications, 2022.

\bibitem{arik2021tabnet}
Sercan~{\"O} Arik and Tomas Pfister.
\newblock Tabnet: Attentive interpretable tabular learning.
\newblock In {\em Proceedings of the AAAI Conference on Artificial Intelligence (AAAI)}, 2021.

\bibitem{arun2016loan}
Kumar Arun, Garg Ishan, and Kaur Sanmeet.
\newblock Loan approval prediction based on machine learning approach.
\newblock {\em IOSR J. Comput. Eng}, 18(3):18--21, 2016.

\bibitem{asuncion2007uci}
Arthur Asuncion and David Newman.
\newblock Uci machine learning repository, 2007.

\bibitem{barocas2023fairness}
Solon Barocas, Moritz Hardt, and Arvind Narayanan.
\newblock {\em Fairness and machine learning: Limitations and opportunities}.
\newblock MIT Press, 2023.

\bibitem{borisov2021deep}
Vadim Borisov, Tobias Leemann, Kathrin Se{\ss}ler, Johannes Haug, Martin Pawelczyk, and Gjergji Kasneci.
\newblock Deep neural networks and tabular data: A survey.
\newblock {\em arXiv preprint arXiv:2110.01889}, 2021.

\bibitem{buczak2015survey}
Anna~L Buczak and Erhan Guven.
\newblock A survey of data mining and machine learning methods for cyber security intrusion detection.
\newblock {\em IEEE Communications surveys \& tutorials}, 18(2):1153--1176, 2015.

\bibitem{chandola2009anomaly}
Varun Chandola, Arindam Banerjee, and Vipin Kumar.
\newblock Anomaly detection: A survey.
\newblock {\em ACM computing surveys (CSUR)}, 41(3):1--58, 2009.

\bibitem{chandrashekar2014survey}
Girish Chandrashekar and Ferat Sahin.
\newblock A survey on feature selection methods.
\newblock {\em Computers \& Electrical Engineering}, 2014.

\bibitem{chen2022danets}
Jintai Chen, Kuanlun Liao, Yao Wan, Danny~Z Chen, and Jian Wu.
\newblock Danets: Deep abstract networks for tabular data classification and regression.
\newblock In {\em Proceedings of the AAAI Conference on Artificial Intelligence (AAAI)}, 2022.

\bibitem{chen2024excelformerneuralnetworksurpassing}
Jintai Chen, Jiahuan Yan, Qiyuan Chen, Danny~Ziyi Chen, Jian Wu, and Jimeng Sun.
\newblock Excelformer: A neural network surpassing gbdts on tabular data, 2024.

\bibitem{chen2016xgboost}
Tianqi Chen and Carlos Guestrin.
\newblock Xgboost: A scalable tree boosting system.
\newblock In {\em Proceedings of the 22nd acm sigkdd international conference on knowledge discovery and data mining}, pages 785--794, 2016.

\bibitem{cherepanova2023performance}
Valeriia Cherepanova, Roman Levin, Gowthami Somepalli, Jonas Geiping, C~Bayan Bruss, Andrew~Gordon Wilson, Tom Goldstein, and Micah Goldblum.
\newblock A performance-driven benchmark for feature selection in tabular deep learning.
\newblock In {\em Thirty-seventh Conference on Neural Information Processing Systems Datasets and Benchmarks Track}, 2023.

\bibitem{clements2020sequential}
Jillian~M Clements, Di~Xu, Nooshin Yousefi, and Dmitry Efimov.
\newblock Sequential deep learning for credit risk monitoring with tabular financial data.
\newblock {\em arXiv preprint arXiv:2012.15330}, 2020.

\bibitem{conover1999practical}
William~Jay Conover.
\newblock {\em Practical nonparametric statistics}, volume 350.
\newblock john wiley \& sons, 1999.

\bibitem{cortes1995support}
Corinna Cortes and Vladimir Vapnik.
\newblock Support-vector networks.
\newblock {\em Machine learning}, 1995.

\bibitem{cover1967nearest}
Thomas Cover and Peter Hart.
\newblock Nearest neighbor pattern classification.
\newblock {\em IEEE transactions on information theory}, 1967.

\bibitem{cox1958regression}
David~R Cox.
\newblock The regression analysis of binary sequences.
\newblock {\em Journal of the Royal Statistical Society: Series B (Methodological)}, 1958.

\bibitem{dettmers2023qlora}
Tim Dettmers, Artidoro Pagnoni, Ari Holtzman, and Luke Zettlemoyer.
\newblock Qlora: Efficient finetuning of quantized llms.
\newblock In {\em Proceedings of the Annual Conference on Neural Information Processing Systems (NeurIPS)}, 2023.

\bibitem{dooley2023forecastpfn}
Samuel Dooley, Gurnoor~Singh Khurana, Chirag Mohapatra, Siddartha~Venkat Naidu, and Colin White.
\newblock Forecastpfn: Synthetically-trained zero-shot forecasting.
\newblock In {\em Proceedings of the Annual Conference on Neural Information Processing Systems (NeurIPS)}, 2023.

\bibitem{dwork2012fairness}
Cynthia Dwork, Moritz Hardt, Toniann Pitassi, Omer Reingold, and Richard Zemel.
\newblock Fairness through awareness.
\newblock In {\em Proceedings of the 3rd innovations in theoretical computer science conference}, 2012.

\bibitem{farthestpointsampling}
Y.~Eldar, M.~Lindenbaum, M.~Porat, and Y.Y. Zeevi.
\newblock The farthest point strategy for progressive image sampling.
\newblock In {\em Proceedings of the 12th IAPR International Conference on Pattern Recognition, Vol. 2 - Conference B: Computer Vision and Image Processing. (Cat. No.94CH3440-5)}, pages 93--97 vol.3, 1994.

\bibitem{feng2022latent}
Leo Feng, Hossein Hajimirsadeghi, Yoshua Bengio, and Mohamed~Osama Ahmed.
\newblock Latent bottlenecked attentive neural processes.
\newblock In {\em The Eleventh International Conference on Learning Representations}, 2023.

\bibitem{feng2023memory}
Leo Feng, Frederick Tung, Hossein Hajimirsadeghi, Yoshua Bengio, and Mohamed~Osama Ahmed.
\newblock Memory efficient neural processes via constant memory attention block.
\newblock {\em OpenReview}, 2023.

\bibitem{friedman2001greedy}
Jerome~H Friedman.
\newblock Greedy function approximation: a gradient boosting machine.
\newblock {\em Annals of statistics}, pages 1189--1232, 2001.

\bibitem{friedman1937use}
Milton Friedman.
\newblock The use of ranks to avoid the assumption of normality implicit in the analysis of variance.
\newblock {\em Journal of the american statistical association}, 1937.

\bibitem{gorishniy2021revisiting}
Yury Gorishniy, Ivan Rubachev, Valentin Khrulkov, and Artem Babenko.
\newblock Revisiting deep learning models for tabular data.
\newblock {\em Proceedings of the Annual Conference on Neural Information Processing Systems (NeurIPS)}, 2021.

\bibitem{grinsztajn2022tree}
Leo Grinsztajn, Edouard Oyallon, and Gael Varoquaux.
\newblock Why do tree-based models still outperform deep learning on typical tabular data?
\newblock In {\em Thirty-sixth Conference on Neural Information Processing Systems Datasets and Benchmarks Track}, 2022.

\bibitem{guo2017deepfm}
Huifeng Guo, Ruiming Tang, Yunming Ye, Zhenguo Li, and Xiuqiang He.
\newblock Deepfm: a factorization-machine based neural network for ctr prediction.
\newblock In {\em IJCAI}, 2017.

\bibitem{guo2022versatile}
Zongyu Guo, Cuiling Lan, Zhizheng Zhang, Yan Lu, and Zhibo Chen.
\newblock Versatile neural processes for learning implicit neural representations.
\newblock In {\em The Eleventh International Conference on Learning Representations}, 2023.

\bibitem{hollmann2022tabpfn}
Noah Hollmann, Samuel M{\"u}ller, Katharina Eggensperger, and Frank Hutter.
\newblock Tabpfn: A transformer that solves small tabular classification problems in a second.
\newblock In {\em Proceedings of the International Conference on Learning Representations (ICLR)}, 2023.

\bibitem{holm1979simple}
Sture Holm.
\newblock A simple sequentially rejective multiple test procedure.
\newblock {\em Scandinavian journal of statistics}, pages 65--70, 1979.

\bibitem{hu2021lora}
Edward~J Hu, Phillip Wallis, Zeyuan Allen-Zhu, Yuanzhi Li, Shean Wang, Lu~Wang, Weizhu Chen, et~al.
\newblock Lora: Low-rank adaptation of large language models.
\newblock In {\em Proceedings of the International Conference on Learning Representations (ICLR)}, 2022.

\bibitem{huang2020tabtransformer}
Xin Huang, Ashish Khetan, Milan Cvitkovic, and Zohar Karnin.
\newblock Tabtransformer: Tabular data modeling using contextual embeddings.
\newblock {\em arXiv preprint arXiv:2012.06678}, 2020.

\bibitem{jiang2024mixtral}
Albert~Q Jiang, Alexandre Sablayrolles, Antoine Roux, Arthur Mensch, Blanche Savary, Chris Bamford, Devendra~Singh Chaplot, Diego de~las Casas, Emma~Bou Hanna, Florian Bressand, et~al.
\newblock Mixtral of experts.
\newblock {\em arXiv preprint arXiv:2401.04088}, 2024.

\bibitem{jiang2020can}
Zhengbao Jiang, Frank~F Xu, Jun Araki, and Graham Neubig.
\newblock How can we know what language models know?
\newblock {\em Transactions of the Association for Computational Linguistics}, 2020.

\bibitem{johnson2016mimic}
Alistair~EW Johnson, Tom~J Pollard, Lu~Shen, Li-wei~H Lehman, Mengling Feng, Mohammad Ghassemi, Benjamin Moody, Peter Szolovits, Leo Anthony~Celi, and Roger~G Mark.
\newblock Mimic-iii, a freely accessible critical care database.
\newblock {\em Scientific data}, 3(1):1--9, 2016.

\bibitem{kadra2021well}
Arlind Kadra, Marius Lindauer, Frank Hutter, and Josif Grabocka.
\newblock Well-tuned simple nets excel on tabular datasets.
\newblock {\em Proceedings of the Annual Conference on Neural Information Processing Systems (NeurIPS)}, 34, 2021.

\bibitem{ke2017lightgbm}
Guolin Ke, Qi~Meng, Thomas Finley, Taifeng Wang, Wei Chen, Weidong Ma, Qiwei Ye, and Tie-Yan Liu.
\newblock Lightgbm: A highly efficient gradient boosting decision tree.
\newblock In {\em Proceedings of the Annual Conference on Neural Information Processing Systems (NeurIPS)}, 2017.

\bibitem{kirichenko2022last}
Polina Kirichenko, Pavel Izmailov, and Andrew~Gordon Wilson.
\newblock Last layer re-training is sufficient for robustness to spurious correlations.
\newblock {\em arXiv preprint arXiv:2204.02937}, 2022.

\bibitem{le2023last}
Phuong~Quynh Le, J{\"o}rg Schl{\"o}tterer, and Christin Seifert.
\newblock Is last layer re-training truly sufficient for robustness to spurious correlations?
\newblock {\em arXiv preprint arXiv:2308.00473}, 2023.

\bibitem{lester2021power}
Brian Lester, Rami Al-Rfou, and Noah Constant.
\newblock The power of scale for parameter-efficient prompt tuning.
\newblock In {\em Proceedings of the 2021 Conference on Empirical Methods in Natural Language Processing}, 2021.

\bibitem{levin2023transfer}
Roman Levin, Valeriia Cherepanova, Avi Schwarzschild, Arpit Bansal, C~Bayan Bruss, Tom Goldstein, Andrew~Gordon Wilson, and Micah Goldblum.
\newblock Transfer learning with deep tabular models.
\newblock {\em ICLR}, 2023.

\bibitem{liaw2002classification}
Andy Liaw, Matthew Wiener, et~al.
\newblock Classification and regression by randomforest.
\newblock {\em R news}, 2(3):18--22, 2002.

\bibitem{liu2023pre}
Pengfei Liu, Weizhe Yuan, Jinlan Fu, Zhengbao Jiang, Hiroaki Hayashi, and Graham Neubig.
\newblock Pre-train, prompt, and predict: A systematic survey of prompting methods in natural language processing.
\newblock {\em ACM Computing Surveys}, 2023.

\bibitem{luo2018neural}
Renqian Luo, Fei Tian, Tao Qin, Enhong Chen, and Tie-Yan Liu.
\newblock Neural architecture optimization.
\newblock In {\em Proceedings of the Annual Conference on Neural Information Processing Systems (NeurIPS)}, 2018.

\bibitem{mcelfresh2023neural}
Duncan McElfresh, Sujay Khandagale, Jonathan Valverde, Ganesh Ramakrishnan, Vishak Prasad, Micah Goldblum, and Colin White.
\newblock When do neural nets outperform boosted trees on tabular data?
\newblock In {\em Proceedings of the Annual Conference on Neural Information Processing Systems (NeurIPS)}, 2023.

\bibitem{mcmahan2013ad}
H~Brendan McMahan, Gary Holt, David Sculley, Michael Young, Dietmar Ebner, Julian Grady, Lan Nie, Todd Phillips, Eugene Davydov, Daniel Golovin, et~al.
\newblock Ad click prediction: a view from the trenches.
\newblock In {\em Proceedings of the Annual Conference on Knowledge Discovery and Data Mining (KDD)}, pages 1222--1230, 2013.

\bibitem{mehrabi2021survey}
Ninareh Mehrabi, Fred Morstatter, Nripsuta Saxena, Kristina Lerman, and Aram Galstyan.
\newblock A survey on bias and fairness in machine learning.
\newblock {\em ACM computing surveys (CSUR)}, 2021.

\bibitem{muller2023pfns4bo}
Samuel M{\"u}ller, Matthias Feurer, Noah Hollmann, and Frank Hutter.
\newblock Pfns4bo: In-context learning for bayesian optimization.
\newblock In {\em Proceedings of the International Conference on Machine Learning (ICML)}, 2023.

\bibitem{muller2022transformers}
Samuel M{\"u}ller, Noah Hollmann, Sebastian~Pineda Arango, Josif Grabocka, and Frank Hutter.
\newblock Transformers can do bayesian inference.
\newblock In {\em Proceedings of the International Conference on Learning Representations (ICLR)}, 2022.

\bibitem{munteanu2018coresets}
Alexander Munteanu and Chris Schwiegelshohn.
\newblock Coresets-methods and history: A theoreticians design pattern for approximation and streaming algorithms.
\newblock {\em KI-K{\"u}nstliche Intelligenz}, 2018.

\bibitem{müller2023mothernet}
Andreas Müller, Carlo Curino, and Raghu Ramakrishnan.
\newblock Mothernet: A foundational hypernetwork for tabular classification, 2023.

\bibitem{nagler2023statistical}
Thomas Nagler.
\newblock Statistical foundations of prior-data fitted networks.
\newblock In {\em Proceedings of the International Conference on Machine Learning (ICML)}, 2023.

\bibitem{pedregosa2011scikit}
Fabian Pedregosa, Ga{\"e}l Varoquaux, Alexandre Gramfort, Vincent Michel, Bertrand Thirion, Olivier Grisel, Mathieu Blondel, Peter Prettenhofer, Ron Weiss, Vincent Dubourg, et~al.
\newblock Scikit-learn: Machine learning in python.
\newblock {\em the Journal of machine Learning research}, 12:2825--2830, 2011.

\bibitem{Popov2020Neural}
Sergei Popov, Stanislav Morozov, and Artem Babenko.
\newblock Neural oblivious decision ensembles for deep learning on tabular data.
\newblock In {\em Proceedings of the International Conference on Learning Representations (ICLR)}, 2020.

\bibitem{prokhorenkova2018catboost}
Liudmila Prokhorenkova, Gleb Gusev, Aleksandr Vorobev, Anna~Veronika Dorogush, and Andrey Gulin.
\newblock Catboost: unbiased boosting with categorical features.
\newblock {\em Proceedings of the Annual Conference on Neural Information Processing Systems (NeurIPS)}, 2018.

\bibitem{quinlan1986induction}
J.~Ross Quinlan.
\newblock Induction of decision trees.
\newblock {\em Machine learning}, 1986.

\bibitem{rastogi2022semi}
Richa Rastogi, Yair Schiff, Alon Hacohen, Zhaozhi Li, Ian Lee, Yuntian Deng, Mert~R Sabuncu, and Volodymyr Kuleshov.
\newblock Semi-parametric inducing point networks and neural processes.
\newblock In {\em The Eleventh International Conference on Learning Representations}, 2023.

\bibitem{richardson2007predicting}
Matthew Richardson, Ewa Dominowska, and Robert Ragno.
\newblock Predicting clicks: estimating the click-through rate for new ads.
\newblock In {\em Proceedings of the 16th international conference on World Wide Web}, pages 521--530, 2007.

\bibitem{rubachev2022revisiting}
Ivan Rubachev, Artem Alekberov, Yury Gorishniy, and Artem Babenko.
\newblock Revisiting pretraining objectives for tabular deep learning.
\newblock {\em arXiv preprint arXiv:2207.03208}, 2022.

\bibitem{savani2020intra}
Yash Savani, Colin White, and Naveen~Sundar Govindarajulu.
\newblock Intra-processing methods for debiasing neural networks.
\newblock {\em Proceedings of the Annual Conference on Neural Information Processing Systems (NeurIPS)}, 2020.

\bibitem{shwartz2022tabular}
Ravid Shwartz-Ziv and Amitai Armon.
\newblock Tabular data: Deep learning is not all you need.
\newblock {\em Information Fusion}, 81:84--90, 2022.

\bibitem{smith1988using}
Jack~W Smith, James~E Everhart, WC~Dickson, William~C Knowler, and Robert~Scott Johannes.
\newblock Using the adap learning algorithm to forecast the onset of diabetes mellitus.
\newblock In {\em Proceedings of the annual symposium on computer application in medical care}, page 261. American Medical Informatics Association, 1988.

\bibitem{somepalli2021saint}
Gowthami Somepalli, Micah Goldblum, Avi Schwarzschild, C~Bayan Bruss, and Tom Goldstein.
\newblock Saint: Improved neural networks for tabular data via row attention and contrastive pre-training.
\newblock {\em arXiv preprint arXiv:2106.01342}, 2021.

\bibitem{tsimpoukelli2021multimodal}
Maria Tsimpoukelli, Jacob~L Menick, Serkan Cabi, SM~Eslami, Oriol Vinyals, and Felix Hill.
\newblock Multimodal few-shot learning with frozen language models.
\newblock {\em Advances in Neural Information Processing Systems}, 34:200--212, 2021.

\bibitem{ulmer2020trust}
Dennis Ulmer, Lotta Meijerink, and Giovanni Cin{\`a}.
\newblock Trust issues: Uncertainty estimation does not enable reliable ood detection on medical tabular data.
\newblock In {\em Machine Learning for Health}, pages 341--354. PMLR, 2020.

\bibitem{urban2021deep}
Christopher~J Urban and Kathleen~M Gates.
\newblock Deep learning: A primer for psychologists.
\newblock {\em Psychological Methods}, 2021.

\bibitem{BLS2023}
{US Bureau of Labor Statistics}.
\newblock National longitudinal surveys of youth data set, 2023.

\bibitem{vanschoren2014openml}
Joaquin Vanschoren, Jan~N Van~Rijn, Bernd Bischl, and Luis Torgo.
\newblock Openml: networked science in machine learning.
\newblock {\em ACM SIGKDD Explorations Newsletter}, 2014.

\bibitem{vergara2014review}
Jorge~R Vergara and Pablo~A Est{\'e}vez.
\newblock A review of feature selection methods based on mutual information.
\newblock {\em Neural computing and applications}, 24:175--186, 2014.

\bibitem{verma2018fairness}
Sahil Verma and Julia Rubin.
\newblock Fairness definitions explained.
\newblock In {\em Proceedings of the international workshop on software fairness}, pages 1--7, 2018.

\bibitem{wallace2019universal}
Eric Wallace, Shi Feng, Nikhil Kandpal, Matt Gardner, and Sameer Singh.
\newblock Universal adversarial triggers for attacking and analyzing nlp.
\newblock In {\em Proceedings of the 2019 Conference on Empirical Methods in Natural Language Processing and the 9th International Joint Conference on Natural Language Processing (EMNLP-IJCNLP)}, 2019.

\bibitem{misc_breast_cancer_wisconsin_(diagnostic)_17}
William Wolberg, Olvi Mangasarian, Nick Street, and W.~Street.
\newblock {Breast Cancer Wisconsin (Diagnostic)}.
\newblock UCI Machine Learning Repository, 1995.
\newblock {DOI}: https://doi.org/10.24432/C5DW2B.

\bibitem{xie2022explanation}
Sang~Michael Xie, Aditi Raghunathan, Percy Liang, and Tengyu Ma.
\newblock An explanation of in-context learning as implicit bayesian inference.
\newblock In {\em International Conference on Learning Representations}, 2022.

\bibitem{yamada2020feature}
Yutaro Yamada, Ofir Lindenbaum, Sahand Negahban, and Yuval Kluger.
\newblock Feature selection using stochastic gates.
\newblock In {\em Proceedings of the International Conference on Machine Learning (ICML)}, 2020.

\bibitem{yoon2020vime}
Jinsung Yoon, Yao Zhang, James Jordon, and Mihaela van~der Schaar.
\newblock Vime: Extending the success of self-and semi-supervised learning to tabular domain.
\newblock {\em Advances in Neural Information Processing Systems}, 33:11033--11043, 2020.

\bibitem{zhao2023survey}
Wayne~Xin Zhao, Kun Zhou, Junyi Li, Tianyi Tang, Xiaolei Wang, Yupeng Hou, Yingqian Min, Beichen Zhang, Junjie Zhang, Zican Dong, et~al.
\newblock A survey of large language models.
\newblock {\em arXiv preprint arXiv:2303.18223}, 2023.

\bibitem{zheng2018fairness}
Yong Zheng, Tanaya Dave, Neha Mishra, and Harshit Kumar.
\newblock Fairness in reciprocal recommendations: A speed-dating study.
\newblock In {\em Adjunct publication of the 26th conference on user modeling, adaptation and personalization}, 2018.

\bibitem{zhong2021factual}
Zexuan Zhong, Dan Friedman, and Danqi Chen.
\newblock Factual probing is [mask]: Learning vs. learning to recall.
\newblock In {\em Proceedings of the 2021 Conference of the North American Chapter of the Association for Computational Linguistics: Human Language Technologies}, 2021.

\bibitem{zhou2022conditional}
Kaiyang Zhou, Jingkang Yang, Chen~Change Loy, and Ziwei Liu.
\newblock Conditional prompt learning for vision-language models.
\newblock In {\em Proceedings of the IEEE/CVF Conference on Computer Vision and Pattern Recognition}, 2022.

\bibitem{zhou2022learning}
Kaiyang Zhou, Jingkang Yang, Chen~Change Loy, and Ziwei Liu.
\newblock Learning to prompt for vision-language models.
\newblock {\em International Journal of Computer Vision}, 2022.

\end{thebibliography}
\bibliographystyle{plain}

\clearpage
\appendix
\section{Broader Societal Impact Statement} \label{sec:impact}

The goal of our work is to investigate context optimization strategies for PFNs, such as prompt-tuning and sketching, which we use to scale and improve the performance of TabPFN.
We do not see any negative broader societal impacts of our work that do not already exist in other classification methods.
In fact, our work may further facilitate the adoption of TabPFN, which has the benefit of being pretrained and therefore having a lower carbon footprint compared to most deep learning approaches that must be trained from scratch.

Furthermore, we demonstrate that our prompt-tuning strategy makes it possible to mitigate the bias of TabPFN while only fine-tuning a small set of embedding vectors; the authors of TabPFN mentioned that it is critical to study and improve TabPFN under the lens of algorithmic fairness and other dimensions of trustworthy AI \citep{hollmann2022tabpfn}.
This may allow practitioners to use TabPFN in sensitive settings for the first time, in which the original TabPFN would lead to biased preditions.
Overall, our hope is that our work will have a positive impact for both practitioners and researchers, by facilitating the adoption of a model with a low carbon footprint, and by providing the tools to mitigate the bias of PFNs.
Likewise, we hope that the possibility to summarize datasets with prompt tuning will add to the toolbox of machine learning practitioners aiming to analyze and interpret their data better, and therefore may have a positive impact on the trustworthieness of machine learning.

\section{Additional Related Work} \label{app:related_work}

\paragraph{Tabular classification.}
Tabular datasets are the oldest and among the most widely used dataset types in machine learning \citep{borisov2021deep,shwartz2022tabular}.
GBDTs \citep{friedman2001greedy} build an ensemble of decision trees, with each tree fitting the residual
of the loss from the previous tree. %, using gradient descent to minimize the losses.
XGBoost \citep{chen2016xgboost} and CatBoost \citep{prokhorenkova2018catboost} are two of the most widely-used and highest-performing GBDTs.
Researchers have also explored many methods based on neural nets \citep{gorishniy2021revisiting,huang2020tabtransformer,kadra2021well}.

There is an active debate in the community on which family of methods perform best on tabular data: neural nets \citep{kadra2021well,arik2021tabnet,Popov2020Neural,rubachev2022revisiting, levin2023transfer}
or GBDTs \citep{shwartz2022tabular,borisov2021deep,gorishniy2021revisiting,grinsztajn2022tree}, with the exception of small datasets, on which TabPFN performs the best \citep{mcelfresh2023neural,hollmann2022tabpfn}.
Finally, sketching and feature selection methods have been extensively studied in prior works  \citep{munteanu2018coresets,chandrashekar2014survey}.

\paragraph{Neural Processes and Prior-Data Fitted Networks.}

Prior-data Fitted Networks (PFNs) \citep{muller2022transformers,hollmann2022tabpfn} are a recently-proposed paradigm for machine learning, which show that fast approximate Bayesian inference is possible by training a neural network to mimic the posterior predictive distribution (PPD) in a single forward pass using in-context learning \citep{muller2022transformers,muller2023pfns4bo,nagler2023statistical}. PFNs were shown to yield state-of-the-art empirical performance on small tabular datasets \citep{hollmann2022tabpfn, mcelfresh2023neural}. PFNs have been used in other applications, including Bayesian optimization \citep{muller2023pfns4bo}, forecasting \citep{dooley2023forecastpfn}, and learning curve extrapolation \citep{adriaensen2022efficient}.

PFNs are neural processes (NPs) \citep{muller2022transformers}. Recent advances in NP are similar in nature to our work, especially recent works that aim to scale attention-based methods to larger contexts.
Feng et al.\ \citep{feng2022latent} propose Latent Bottlenecked Attentive Neural Processes (LBANPs), a transformer-based neural process which overcomes the quadratic complexity of transformers by encoding the context dataset into a constant number of latent vectors.
Guo et al.\ \citep{guo2022versatile} propose Versatile Neural Processes (VNP), which increases the capability of NPs to handle compex signals by using a new bottleneck encoder.
\citep{rastogi2022semi} introduces semi-parametric inducing point networks (SPIN), which can attend to a training set at inference time with linear complexity via inducing point methods.
\citep{feng2023memory} introduce Constant Memory Attention Block (CMAB), an attention block that is permutation-invariant and has constant memory complexity when computing the output, as well as  Constant Memory Attentive Neural Processes (CMANPs), a NP that requires constant memory.

For additional related work, see \cref{app:related_work}.

\paragraph{Prompt tuning.}
Researchers have extensively studied prompting techniques for large language models (LLMs) \citep{liu2023pre}.
%Unlike traditional supervised learning, in prompt-based learning, LLMs model the probability of output text directly. 
In such methods, one modifies the input into a \emph{prompt}, and the LLM predicts the subsequent tokens, from which it derives the output predictions.
\emph{Prompt tuning} techniques typically start with a pretrained LLM and use the training data of a downstream task to find the best prompt for this task. 
`Hard' prompt tuning involves finding the best natural language prompts using discrete search techniques \citep{jiang2020can,wallace2019universal}, while `soft' prompt tuning optimizes a prompt directly in the embedding space of the model \citep{lester2021power,zhong2021factual}. Soft prompt tuning has also been applied to multi-modal models such as vision-language models \citep{zhou2022learning,zhou2022conditional,tsimpoukelli2021multimodal}.

\section{Additional experimental details} \label{app:experiments}

\begin{table}[t]
\caption{List of all datasets used in \cref{sec:experiments} and \cref{app:experiments}. \newbenchmark datasets are listed in part one. Datasets in part two were used for the class extension experiments in \cref{tab:large_cardinality}. 
Datasets in part three were used for fairness experiments in \cref{sec:fairness}.
The additional dataset in part four was used for intepretability experiments in \cref{sec:fairness}.
Datasets in part 5 were used for the ablations on sketching and feature selection in \cref{app:experiments}.
For the list of \ndatasets{} used in \cref{tab:full_results}, see
\url{https://github.com/penfever/TuneTables/blob/main/datasets-used.csv}.
}
\label{tab:datasets}
\centering
\resizebox{.7\textwidth}{!}{%
\begin{tabular}{lrrr}
\\
\toprule
Dataset & num.\ classes & num.\ features & num.\ samples \\
\midrule
click-prediction-small & 2 & 11 & 1997410 \\
poker-hand & 10 & 10 & 1025009 \\
agrawal1 & 2 & 9 & 1000000 \\
BNG (labor) & 2 & 16 & 1000000 \\
airlines & 2 & 7 & 539383 \\
albert & 2 & 78 & 425240 \\
BNG (vote) & 2 & 16 & 131072 \\
connect-4 & 2 & 28 & 98050 \\
higgs & 3 & 42 & 67557 \\
volkert & 10 & 180 & 58310 \\
riccardo & 2 & 4296 & 20000 \\
elevators & 2 & 18 & 16599 \\
eeg-eye-state & 2 & 14 & 14980 \\
har & 6 & 561 & 10299 \\
dilbert & 5 & 2000 & 10000 \\
robert & 10 & 7200 & 10000 \\
christine & 2 & 1636 & 5418 \\
bioresponse & 2 & 1776 & 3151 \\
kc-1 & 2 & 21 & 2109 \\
car & 4 & 6 & 1728 \\
cmc & 3 & 9 & 1473 \\
blood-transfusion & 2 & 4 & 748 \\
balance-scale & 3 & 4 & 625 \\
climate & 2 & 18 & 540 \\
cylinder-bands & 2 & 37 & 540 \\
dresses-sales & 2 & 12 & 500 \\
colic & 2 & 22 & 368 \\
ecoli & 8 & 7 & 336 \\
heart-c & 2 & 13 & 303 \\
breast-cancer & 2 & 9 & 286 \\
\midrule
openml\_\_ldpa\_\_9974 & 11 & 7 & 164860 \\
openml\_\_walking-activity\_\_9945 & 22 & 4 & 149332 \\
openml\_\_Devnagari-Script\_\_167121 & 46 & 1024 & 92000 \\
openml\_\_helena\_\_168329 & 100 & 27 & 65196 \\
openml\_\_chess\_\_3952 & 18 & 6 & 28056 \\
openml\_\_kropt\_\_2076 & 18 & 6 & 28056 \\
openml\_\_letter\_\_6 & 26 & 16 & 20000 \\
openml\_\_isolet\_\_3481 & 26 & 617 & 7797 \\
openml\_\_texture\_\_125922 & 11 & 40 & 5500 \\
openml\_\_one-hundred-plants-texture\_\_9956 & 100 & 64 & 1599 \\
openml\_\_vowel\_\_3022 & 11 & 12 & 990 \\
openml\_\_soybean\_\_41 & 19 & 35 & 683 \\
openml\_\_collins\_\_3567 & 15 & 21 & 500 \\
openml\_\_arrhythmia\_\_5 & 13 & 279 & 452 \\
openml\_\_primary-tumor\_\_146032 & 21 & 17 & 339 \\
\midrule
% Valeriia's datasets
adult-census & 2 & 15 & 32561 \\
SpeedDating &  2 & 121 & 8378\\
compas-two-years & 2 &  12 & 4966 \\
nlsy (national-longitudinal-survey-binary) & 2 & 17 & 4908\\
\midrule
Diabetes & 322 & 10 & 442 \\
\midrule
skin-segmentation & 2 & 3 & 245057 \\
FM (Fashion-MNIST) & 10 & 784 & 70000 \\
CIFAR-10 & 10 & 3072 & 60000 \\
gddc (gas-drift-different-concentrations) & 6 & 129 & 13910 \\
pendigits & 10 & 16 & 10992 \\
mfeat-factors & 10 & 216 & 2000 \\
mfeat-pixel & 10 & 240 & 2000 \\
semeion & 10 & 256 & 1593 \\
hill-valley & 2 & 100 & 1212 \\
\bottomrule
\end{tabular}
}
\end{table}

\paragraph{Hyperparameter settings for baselines.} 
For the baselines, we use the same hyperparameter search space as in prior work \citep{mcelfresh2023neural} (which is itself similar to other works \citep{kadra2021well}). We choose the best configuration among the default parameter sets and the additional 29 sets, using k-fold cross-validation with 10 folds. We follow the best configurations of \citep{mcelfresh2023neural} with exceptions; for CatBoost and XGBoost, we increase the range of tree depth by a factor of 10, since a high depth is often more appropriate for very large datasets. We find that this improves performance on large datasets, but sometimes harms performance on small datasets, and increases the runtime -- see \cref{app:deepgbdt} for the raw results. For each dataset, each algorithm is allowed a maximum runtime of 10 hours for the entire search process. For ExcelFormer, we decrease the number of dimensions and depth of the default transformer, as we find that without this modification, it times out on even medium-sized datasets.

\paragraph{Hyperparameter settings for TuneTables.}   Consistent with the experimental design in \cite{mcelfresh2023neural}, we perform  hyperparameter optimization of TuneTables for up to 30 variants; however, we utilize a grid search rather than Optuna. We use a batch size of $1$ and no gradient aggregation.  The space over which we conduct grid search in TuneTables is conditioned on metadata; number of features, number of  samples, number of classes. The feature and class splits are set by the limits of the particular TabPFN checkpoint we optimize. When we reach a leaf node, TuneTables-standard and TuneTables-medium conduct a grid search over a fixed range of configurations. The size of our search space is always < 30, and usually < 10. For feature-large datasets, TuneTables-light conducts additional optimization in the feature space prior to grid search. Further details on our search algorithm can be found in the repository associated with the paper. TuneTables hyperparameter settings can be found in  chart form in \Cref{tab:tthparams}.

\begin{table}
\centering
\caption{TuneTables and TabPFNs3000 hyperparameter configurations based on number of samples.}
\resizebox{\textwidth}{!}{
\begin{tabular}{|c|c|c|c|}
\hline
\textbf{Parameter} & \textbf{TabPFN} & \textbf{TuneTables $\leq$ 2000} & \textbf{TuneTables $>$ 2000} \\ \hline
Batch Size & 1 & - & - \\ \hline
Real Data Qty & 3000 & 0 & \{1152, 0\} \\ \hline
Ensemble Size & 32 & 1 & 10 \\ \hline
Tuned Prompt Dimension & - & 10 & 1000 \\ \hline
Ensemble Method & - & - & Avg Top 2 \\ \hline
Epochs & - & \{7, 60\} & 100 \\ \hline
Warmup & - & 10 & 10 \\ \hline
Sequence Length Per Batch & - & Fixed & \{Variable, Fixed\} \\ \hline
Early Stopping & - & \{2, 6\} & 6 \\ \hline
Learning Rate & - & 3e-2 & 1e-3 \\ \hline
Validation Frequency & - & 2 & 2 \\ \hline
Max Val Size During Training & - & 2000 & 2000 \\ \hline
Optimizer & - & \begin{tabular}[c]{@{}c@{}}AdamW, Default Settings,\\ Weight Decay 0\end{tabular} & \begin{tabular}[c]{@{}c@{}}AdamW, Default Settings,\\ Weight Decay 0\end{tabular} \\ \hline
Loss & - & \begin{tabular}[c]{@{}c@{}} \{Cross-Entropy,\\ KL Divergence\}\end{tabular} & - \\ \hline
Tuned Prompt Labels & - & \{Equal, Proportional\} & - \\ \hline
\end{tabular}}
\label{tab:tthparams}
\end{table}

\subsection{Memory Efficiency} In our experiments, we find that we exceed hardware limitations around 3000 samples with a batch size of 1 when using TabPFN on a GPU with 48GB of VRAM. While FlashAttention, a popular optimization, is memory-linear in sequence length, but it has not yet been implemented in the public release of TabPFN, which relies on an older version of PyTorch. Flash attention may be integrated into a future release, in which case any benefits will carry over to our method. Even so,  inference time continues to be quadratic, and there are other overheads on GPU memory. Overall, there remains a considerable need for new algorithmic methods which can be scaled up to the sequence lengths required by large tabular datasets.

\subsection{Feature selection and sketching} \label{app:sketching}

We ablate strategies for both sketching (subsampling) and feature selection for scaling TabPFN.
We consider three sketching methods (\emph{random}, $k$-\emph{means}, and \emph{CoreSet}) and three feature selection methods (\emph{random}, \emph{PCA}, \emph{mutual information}) as described in \cref{sec:background}.
In addition to sketching, we consider two different methods for sampling class labels: \emph{equal} and \emph{proportional}.

For efficient coreset selection, we use a variant of Farthest Point Sampling~\cite{farthestpointsampling}; after selecting an initial set of n=5 random points, we compute the distance of each point in the dataset to the set of already selected points. Next, we add to the set of selected points the point whose distance to the selected points is maximal. Finally, we update the distances of all the points according to the updated selected set; and continue iteratively.

We limit our algorithmic comparison to TabPFN to CatBoost, which is the overall best-performing model in \cite{mcelfresh2023neural}.
We compare all combinations of sketching and feature selection with both CatBoost and TabPFN; see \cref{tab:sketching_fs}.
Interestingly, we find that \emph{random} sketching performs just as well as the more involved algorithms, $k$-means and \emph{CoreSet}. On the other hand, PCA significantly outperforms mutual information and random feature selection methods, when the original number of features is large.

\begin{table*}[t]
\caption{\textbf{Comparative performance of TabPFN and CatBoost with sketching, feature selection, and sampling methods. }  On a distinct subset of the datasets in \cite{mcelfresh2023neural} selected to emphasize datasets with many features or many samples, we compare CatBoost and TabPFNs3000. When both models are limited to 3000 samples, TabPFNs3000 performs better on 12 of 17 datasets where significant differences exist. When CatBoost is allowed access to the entire training data, the win rate is identical. In most cases, random sample selection is sufficient for optimal performance. Both models benefit from PCA and mutual information dimension reduction when the feature space is large. The columns labeled SKT / FTS / SMP list the best performing method for sketching, feature subsampling and label-aware sketching technique, respectively. Label-aware sketching refers to a strategy where we either sample instances proportionate to their labels, or we oversample minority classes with replacement to create a class-balanced distribution. While the choice of label-aware sketching strategy is often impactful (and we use it in TuneTables), and the choice of feature subselection method can be important for some datasets, in all but one case, no sketching method we test outperforms random sampling.
 \textbf{Bold} indicates the best-performing model(s).}
\resizebox{\textwidth}{!}{%
\begin{tabular}{@{}lrrrrrll@{}}
\toprule
 & \multicolumn{1}{l}{\textbf{Full}} & \multicolumn{1}{l}{\textbf{Best}} & \multicolumn{1}{l}{\textbf{Random}} & \multicolumn{1}{l}{\textbf{Best}} & \multicolumn{1}{l}{\textbf{Random}} & \textbf{SKT / FTS / SMP} & \textbf{SKT / FTS / SMP} \\ 
 & CatBoost & CatBoost & CatBoost & TabPFN & TabPFN & CatBoost & TabPFN \\
\midrule
airlines\_189354 & \textbf{0.653} & 0.637 & 0.637 & 0.594 & 0.589 & RND / RND / PR & RND / RND / PR \\
albert\_189356 & \textbf{0.698} & 0.657 & 0.657 & 0.64 & 0.64 & RND / RND / PR & RND / RND / PR \\
CIFAR\_10\_167124 & \textbf{0.434} & 0.37 & 0.342 & 0.373 & 0.372 & RND / PCA / PR & RND / RND / PR \\
connect-4\_146195 & \textbf{0.749} & 0.716 & 0.716 & 0.66 & 0.659 & RND / RND / PR & RND / RND / PR \\
eeg-eye-state\_14951 & 0.832 & 0.808 & 0.806 & \textbf{0.932} & \textbf{0.932} & RND / RND / PR & RND / RND / EQ \\
elevators\_3711 & 0.855 & 0.838 & 0.838 & \textbf{0.9} & \textbf{0.899} & RND / MUT / PR & RND / RND / PR \\
FM\_146825 & \textbf{0.843} & 0.787 & 0.787 & \textbf{0.835} & 0.812 & RND / RND / PR & RND / PCA / PR \\
gddc\_9987 & 0.97 & 0.976 & 0.955 & \textbf{0.994} & \textbf{0.993} & RND / PCA / EQ & RND / RND / PR \\
higgs\_146606 & \textbf{0.71} & 0.684 & 0.684 & 0.665 & 0.661 & RND / RND / PR & RND / RND / PR \\
hill-valley\_145847 & 0.514 & 0.514 & 0.514 & \textbf{0.56} & \textbf{0.56} & RND / RND / PR & RND / RND / PR \\
mfeat-factors\_12 & 0.954 & 0.95 & 0.943 & \textbf{0.973} & \textbf{0.973} & KMN / RND / EQ & RND / RND / PR \\
mfeat-pixel\_146824 & 0.955 & 0.951 & 0.951 & \textbf{0.971} & \textbf{0.97} & RND / RND / PR & RND / RND / PR \\
pendigits\_32 & 0.972 & 0.966 & 0.964 & \textbf{0.995} & \textbf{0.993} & RND / RND / PR & RND / RND / PR \\
poker-hand\_9890 & \textbf{0.664} & 0.572 & 0.561 & 0.519 & 0.515 & RND / RND / PR & RND / RND / PR \\
riccardo\_168338 & 0.951 & 0.956 & 0.93 & \textbf{0.991} & 0.982 & RND / PCA / EQ & RND / MUT / EQ \\
robert\_168332 & \textbf{0.446} & 0.367 & 0.367 & 0.384 & 0.359 & RND / RND / PR & RND / PCA / EQ \\
semeion\_9964 & 0.887 & 0.869 & 0.863 & \textbf{0.915} & \textbf{0.915} & RND / MUT / EQ & RND / RND / PR \\
ss\_9965 & \textbf{0.994} & 0.989 & 0.987 & \textbf{0.999} & \textbf{0.999} & RND / RND / PR & RND / RND / PR \\
volkert\_168331 & \textbf{0.608} & 0.56 & 0.56 & 0.557 & 0.555 & RND / RND / PR & RND / RND / PR \\ 
\midrule
Average & 0.773 &	0.746 &	0.740	& 0.761	& 0.757 & RND / RND / PR & RND / RND / PR \\

\bottomrule
\end{tabular}%
}
\label{tab:sketching_fs}
%\vspace{-4mm}
\end{table*}

\subsection{Additional results}

First, we present the large class table from \cref{sec:experiments}. See \cref{tab:large_cardinality}.

\begin{figure*}[t]
    \centering
    \includegraphics[width=0.7\textwidth]{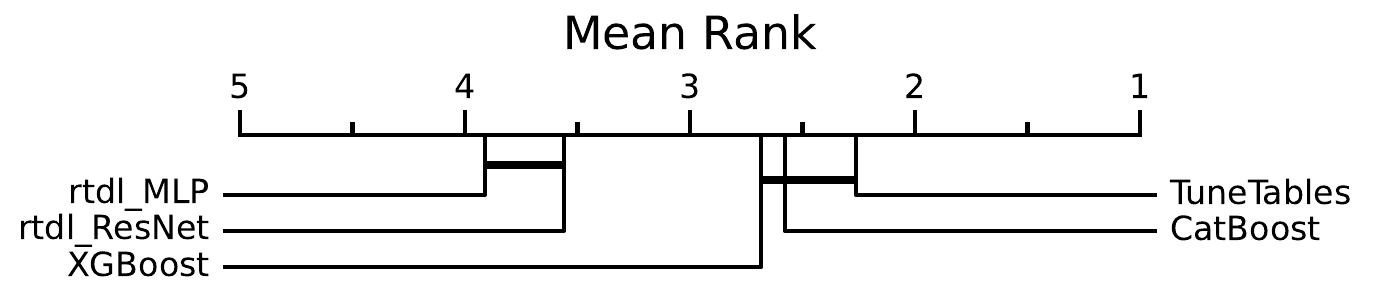}
    \caption{\textbf{TuneTables is competitive with state-of-the-art tabular models.} A critical difference plot according to mean accuracy rank across all \newbenchmark datasets with fewer than 50\,000 samples. Algorithms which are \emph{not significantly different} ($p>0.05$) are connected with a horizontal black bar. TuneTables achieves the highest mean rank of any algorithm. 
    This plot is similar to \cref{fig:critical_difference}, but the search spaces for XGBoost and CatBoost are expanded to include more trees.
    }
    \label{fig:critical_difference_gdbt_var}
\end{figure*}

\begin{table*}[t]
\caption{\textbf{Comparison of algorithms on datasets with a large number of classes.}
TuneTables can effectively handle datasets with more classes than the ones used for pretraining, which was not possible with TabPFN. For each algorithm, we compute its mean test accuracy, and mean rank in terms of accuracy.
We also compute the mean Z-score, computed by normalizing the set of results on each dataset (by mean 0 std.\ 1), so that each dataset has the same weight, and averaging each algorithm's normalized performances. We see that TuneTables performs the best across all performance-oriented metrics. Fractional num.\ wins values are averaged over three splits per dataset, and reflect the presence of multi-way ties on certain splits.
}
\label{tab:large_cardinality}
\centering
\resizebox{.8\textwidth}{!}{%
\begin{tabular}{lrrrrrrr}
\toprule
Method & Mean Acc.\  & Mean Rank & Mean Z-Score & Std.\ Z-Score & Med.\ Z-Score & Num.\ Wins \\
\midrule
XGBoost       & 0.885 & 2.000 & 1.052 & 0.558 & 1.245 & 12.0 \\
TuneTables    & 0.779 & 2.524 & 0.602 & 0.932 & 0.971 & 5.0 \\
rtdl\_ResNet  & 0.776 & 3.929 & 0.053 & 0.826 & -0.006 & 2.0 \\
KNN           & 0.789 & 3.976 & 0.121 & 0.262 & 0.103 & 0.0 \\
CatBoost      & 0.781 & 4.524 & -0.177 & 0.960 & 0.107 & 2.0 \\
RandomForest  & 0.751 & 5.214 & -0.628 & 0.754 & -0.767 & 0.0 \\
rtdl\_MLP     & 0.624 & 5.833 & -1.022 & 0.908 & -1.257 & 0.0 \\
\bottomrule
\end{tabular}
}
\end{table*}

\begin{figure*}[t]
    \centering
    \includegraphics[width=0.7\textwidth]{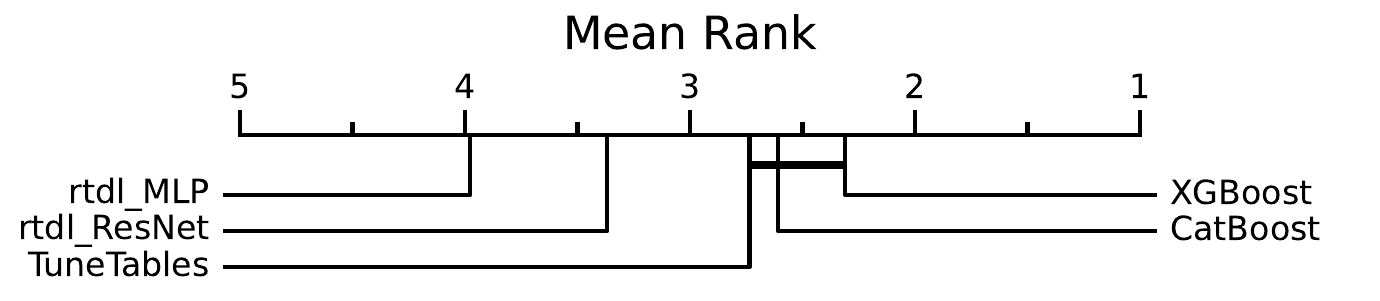}
    \caption{We present a critical difference plot according to mean accuracy rank on \newbenchmark. Algorithms which are \emph{not significantly different} ($p>0.05$) are connected with a horizontal black bar. Across these datasets, TuneTables achieves performance that is not significantly different from CatBoost or XGBoost. 
    }
    \label{fig:critical_difference_all_datasets}
\end{figure*}

In \cref{tab:less_than_50k}, we gave aggregate statistics for TuneTables and baselines across all datasets of size less than 50\,000. Now, in \cref{tab:full_results}, we present results for TuneTables, CatBoost, and XGBoost on all datasets individually, including datasets of size nearly two million.
Note that we report the end-to-end runtime of all methods, including 30 iterations of search (for example, TabPFNs3000 takes about $30\times$ more time than TabPFN as reported in prior work \citep{mcelfresh2023neural}). 
In \cref{fig:critical_difference_all_datasets}, we present a critical difference diagram similar to \cref{fig:critical_difference} but over all 29 datasets.
Across these 29 datasets, TuneTables achieves performance that is not statistically different from CatBoost or XGBoost.

\begin{table}[t]
\caption{
\textbf{Comparison of the top-performing methods on \newbenchmark.} For each algorithm, we show its test accuracy and end-to-end runtime in seconds. In this table, \cref{tab:runtimes_full} and \cref{tab:full_results_gbdt_var}, we report the summed runtime for all random seeds (30 in the case of all algorithms except for TuneTables, which generally requires fewer than 30). Since TuneTables uses TabPFN predictions on small datasets, we conservatively report its end to end runtime as the sum of TabPFNs3000 runtime and TuneTables grid search runtime. This prediction is also extremely conservative for TabPFNs3000, as 30 random seeds are unnecessary on datasets with fewer than 100 features. We report runtime in this manner to mitigate concerns that we are advantaging PFNs over boosted trees. We also show the number of samples in each dataset.
All datasets above the line were included in \cref{tab:less_than_50k}.
}
\label{tab:full_results}
\resizebox{\textwidth}{!}{%
\centering
\begin{tabular}{lrrrrrrrrr}
\toprule
Dataset & Size & \multicolumn{2}{c}{TabPFNs3000} & \multicolumn{2}{c}{TuneTables} & \multicolumn{2}{c}{CatBoost} & \multicolumn{2}{c}{XGBoost} \\
& & Acc.\ & Runtime & Acc.\ & Runtime  & Acc.\ & Runtime  & Acc.\ & Runtime \\
\midrule
breast-cancer & 286 & 0.765 & 29 & 0.770 & 65 & 0.770 & 37 & 0.724 & 4 \\
heart-c & 303 & 0.848 & 40 & 0.903 & 66 & 0.903 & 21 & 0.839 & 2 \\
ecoli & 336 & 0.848 & 30 & 0.843 & 66 & 0.892 & 73 & 0.833 & 17 \\
colic & 368 & 0.856 & 39 & 0.892 & 66 & 0.874 & 218 & 0.883 & 3 \\
dresses-sales & 500 & 0.578 & 41 & 0.580 & 122 & 0.640 & 40 & 0.640 & 5 \\
cylinder-bands & 540 & 0.800 & 41 & 0.846 & 82 & 0.827 & 94 & 0.864 & 6 \\
climate & 540 & 0.959 & 59 & 0.951 & 97 & 0.963 & 25 & 0.926 & 4 \\
balance-scale & 625 & 0.990 & 29 & 0.995 & 55 & 0.899 & 53 & 0.894 & 145 \\
blood-transfusion & 748 & 0.801 & 25 & 0.782 & 56 & 0.756 & 24 & 0.760 & 58 \\
cmc & 1473 & 0.554 & 91 & 0.556 & 109 & 0.561 & 78 & 0.561 & 13 \\
kc-1 & 2109 & 0.862 & 168 & 0.856 & 187 & 0.856 & 49 & 0.859 & 6 \\
bioresponse & 3151 & 0.797 & 638 & 0.798 & 3012 & 0.788 & 113 & 0.800 & 102 \\
christine & 5418 & 0.742 & 666 & 0.755 & 3920 & 0.736 & 331 & 0.743 & 4 \\
robert & 10000 & 0.250 & 964 & 0.414 & 2397 & 0.464 & 599 & 0.503 & 2118 \\
dilbert & 10000 & 0.922 & 761 & 0.992 & 3749 & 0.949 & 809 & 0.978 & 1144 \\
har & 10299 & 0.936 & 370 & 0.981 & 2657 & 0.985 & 302 & 0.994 & 232 \\
eeg-eye-state & 14980 & 0.940 & 178 & 0.986 & 1929 & 0.907 & 28 & 0.946 & 18 \\
elevators & 16599 & 0.902 & 186 & 0.902 & 1297 & 0.891 & 176 & 0.896 & 234 \\
riccardo & 20000 & 0.922 & 1395 & 0.995 & 5247 & 0.997 & 692 & 0.998 & 735 \\

\midrule

volkert & 58310 & 0.567 & 459 & 0.693 & 6331 & 0.666 & 230 & 0.703 & 604 \\
higgs & 67557 & 0.671 & 931 & 0.714 & 4084 & 0.724 & 80 & 0.725 & 46 \\
connect-4 & 98050 & 0.668 & 696 & 0.817 & 5395 & 0.807 & 204 & 0.855 & 106 \\
BNG (vote) & 131072 & 0.968 & 1976 & 0.974 & 2493 & 0.975 & 42 & 0.975 & 16 \\
albert & 425240 & 0.642 & 2363 & 0.658 & 17518 & 0.706 & 1078 & 0.690 & 76 \\
airlines & 539383 & 0.600 & 2602 & 0.653 & 44434 & 0.664 & 155 & 0.673 & 184 \\
BNG (labor) & 1000000 & 0.937 & 5518 & 0.967 & 7717 & 0.971 & 226 & 0.970 & 179 \\
agrawal1 & 1000000 & 0.948 & 5158 & 0.950 & 45504 & 0.951 & 142 & 0.951 & 98 \\
poker-hand & 1025009 & 0.531 & 2423 & 1.000 & 10471 & 0.912 & 5331 & 0.940 & 439 \\
click-prediction-small & 1997410 & 0.833 & 10421 & 0.837 & 33148 & 0.842 & 85 & 0.843 & 213 \\
\midrule
Average & & 0.781	&  1320	&  0.830	&  6975	 & 0.823	& 390 &	0.826 &	234 \\
\bottomrule
\end{tabular}
}
\end{table}

\paragraph{Neural net comparison.}
In \cref{tab:less_than_50k} and \cref{tab:full_results}, we compared TuneTables to GBDTs and two other neural nets. Now, we compare TuneTables to two additional neural nets: MLP and TabTransformer \citep{huang2020tabtransformer} (for four total neural net baselines).
Since transformer-based methods have a higher memory consumption, we obtained full results on 17 total datasets. 
See \cref{tab:neuralnets}.
We find that TuneTables substantially outperforms all other neural nets on average across all datasets, achieving the lowest average rank at 1.93, and achieving the highest average Z-Score of 0.85, far above the second-place neural net's value of 0.18.

\begin{table}[t]
\centering
\caption{\textbf{Comparison of neural nets on \newbenchmark.} We compare TuneTables to other prominent deep learning methods for tabular data on the 17 datasets in \newbenchmark for which all algorithms reported results. For each algorithm, we compute its different metrics of accuracy and rank.
We also compute the mean Z-score, computed by normalizing the set of results on each dataset (by mean 0 std.\ 1), so that each dataset has the same weight, and averaging each algorithm's normalized performances. We see that TuneTables performs the best across all performance-oriented metrics. Fractional num.\ wins values are averaged over three splits per dataset, and reflect the presence of multi-way ties on certain splits.
}
\label{tab:neuralnets}
\resizebox{\textwidth}{!}{%
\begin{tabular}{lrrrrrrr}
\toprule
Method & Mean Acc.\ & Med Acc.\ & Mean Rank & Med.\ Rank & Mean Z-Score & Med.\ Z-Score & Num.\ Wins \\
\midrule
TuneTables     & \textbf{0.81} & \textbf{0.85} & \textbf{1.93} & \textbf{1.0} & \textbf{0.85}  & \textbf{0.91}  & \textbf{9.67} \\
rtdl\_ResNet   & 0.77 & 0.78 & 2.68 & 2.0 & 0.18  & 0.19  & 3.33 \\
TabTransformer& 0.76 & 0.73 & 3.05 & 3.0 & -0.10 & 0.05  & 3.00 \\
rtdl\_MLP      & 0.73 & 0.76 & 3.55 & 4.0 & -0.34 & -0.55 & 0.83 \\
MLP           & 0.72 & 0.74 & 3.79 & 4.0 & -0.59 & -0.55 & 0.17 \\
\bottomrule
\end{tabular}
}
\end{table}

\paragraph{Inference Time}
We give the full details for inference time; see \cref{tab:inference}.

\begin{table}[t]
\caption{
\textbf{Comparison of inference times across datasets.} In this table, we compare the inference time of TabPFNs3000 (which uses up to 3000 points of real data as context, when available) and four common configurations for TuneTables. We report inference time for the entire test set, as well as per 1000 samples, estimated from the inference time for the entire test set. At inference time, TuneTables does not require ensembling to achieve high accuracy, while TabPFNs3000 does. Therefore, TuneTables is considerably faster.
}
\label{tab:inference}
\resizebox{\textwidth}{!}{%
\centering
\begin{tabular}{lrrrrrrrrrrrrrrr}
\toprule
\textbf{Dataset} & \textbf{N.\ classes} & \textbf{N.\ Feats} & \textbf{N.\ Samples} & \multicolumn{2}{l}{\textbf{ TabPFNs3000}} & \multicolumn{2}{l}{\textbf{TuneTables-pt10-C}} & \multicolumn{2}{l}{\textbf{TuneTables-pt10-NC}} & \multicolumn{2}{l}{\textbf{TuneTables-pt1000-C}} & \multicolumn{2}{l}{\textbf{TuneTables-pt1000-NC}} \\
& & & & \textbf{3000 pts, ens.} & \textbf{per1k} & \textbf{3000 pts} & \textbf{per1k} & \textbf{0 pts} & \textbf{per1k} & \textbf{3000 pts} & \textbf{per1k} & \textbf{0 pts} & \textbf{per1k} \\ 
\midrule
breast-cancer & 2 & 9 & 286 & 0.951 & 3.325 & 0.103 & 0.36 & 0.103 & 0.36 & 0.119 & 0.416 & 0.109 & 0.381 \\
heart-c & 2 & 13 & 303 & 1.011 & 3.337 & 0.103 & 0.34 & 0.102 & 0.337 & 0.111 & 0.366 & 0.107 & 0.353 \\
ecoli & 8 & 7 & 336 & 0.99 & 2.946 & 0.094 & 0.28 & 0.092 & 0.274 & 0.094 & 0.28 & 0.091 & 0.271 \\
colic & 2 & 22 & 368 & 1.045 & 2.84 & 0.103 & 0.28 & 0.102 & 0.277 & 0.124 & 0.337 & 0.112 & 0.304 \\
dresses-sales & 2 & 12 & 500 & 1.024 & 2.048 & 0.098 & 0.196 & 0.09 & 0.18 & 0.106 & 0.212 & 0.092 & 0.184 \\
climate & 2 & 18 & 540 & 1.067 & 1.976 & 0.098 & 0.181 & 0.09 & 0.167 & 0.11 & 0.204 & 0.095 & 0.176 \\
cylinder-bands & 2 & 37 & 540 & 1.317 & 2.439 & 0.088 & 0.163 & 0.091 & 0.169 & 0.108 & 0.2 & 0.093 & 0.172 \\
balance-scale & 3 & 4 & 625 & 0.7 & 1.12 & 0.115 & 0.184 & 0.105 & 0.168 & 0.147 & 0.235 & 0.104 & 0.166 \\
blood-transfusion & 2 & 4 & 748 & 0.595 & 0.795 & 0.123 & 0.164 & 0.117 & 0.156 & 0.149 & 0.199 & 0.107 & 0.143 \\
cmc & 3 & 9 & 1473 & 1.485 & 1.008 & 0.099 & 0.067 & 0.093 & 0.063 & 0.155 & 0.105 & 0.093 & 0.063 \\
kc-1 & 2 & 21 & 2109 & 2.623 & 1.244 & 0.12 & 0.057 & 0.11 & 0.052 & 0.203 & 0.096 & 0.116 & 0.055 \\
bioresponse & 2 & 1776 & 3151 & 5.623 & 1.785 & 0.23 & 0.073 & 0.116 & 0.037 & 0.338 & 0.107 & 0.114 & 0.036 \\
christine & 2 & 1636 & 5418 & 5.392 & 0.995 & 0.486 & 0.09 & 0.183 & 0.034 & 0.725 & 0.134 & 0.167 & 0.031 \\
dilbert & 5 & 2000 & 10000 & 9.355 & 0.936 & 0.74 & 0.074 & 0.204 & 0.02 & 1.126 & 0.113 & 0.232 & 0.023 \\
robert & 10 & 7200 & 10000 & 9.369 & 0.937 & 0.718 & 0.072 & 0.202 & 0.02 & 1.143 & 0.114 & 0.208 & 0.021 \\
har&6&561&10299&9.399&0.913&0.804&0.078&0.219&0.021&1.27&0.123&0.236&0.023\\
eeg-eye-state&2&14&14980&8.235&0.55&1.086&0.072&0.245&0.016&1.683&0.112&0.284&0.019\\
elevators&2&18&16599&8.501&0.512&1.169&0.07&0.278&0.017&1.799&0.108&0.294&0.018\\
riccardo&2&4296&20000&5.994&0.3&1.369&0.111&0.325&0.018&2.21&0.111&0.358&0.018\\
volkert&10&180&58310&17.448&0.299&3.931&0.067&0.772&0.013&6.336&0.109&0.897&0.015\\
higgs&3&42&67557&21.991&0.326&6.449&0.095&1.356&0.02&10.404&0.154&1.517&0.022\\
connect-4&2&28&98050&19.966&0.204&4.491&0.046&0.922&0.009&7.064&0.072&0.992&0.01\\
BNG(vote)&2&16&131072&28.229&0.215&8.898&0.068&1.636&0.012&14.25&0.109&1.95&0.015\\
albert&2&78&425240&77.546&0.182&27.988&0.066&5.748&0.014&47.855&0.113&6.378&0.015\\
airlines&2&7&539383&85.146&0.158&36.797&0.068&6.751&0.013&61.477&0.114&8.123&0.015\\
agrawal1&2&9&1000000&193.083&0.193&70.665&0.071&12.129&0.012&112.533&0.113&14.944&0.015\\
BNG(labor)&2&16&1000000&190.81&0.191&69.374&0.069&12.094&0.012&108.211&0.108&14.666&0.015\\
poker-hand&10&10&1025009&198.151&0.193&71.605&0.07&13.097&0.013&112.214&0.109&15.199&0.015\\
click-prediction-small&2&11&1997410&384.192&0.192&133.212&0.067&24.964&0.012&227.546&0.114&30.982&0.016\\
Average&&&222079 &44.525&1.109&15.212&0.124&2.839&0.087&24.814&0.162&3.402&0.09\\
\bottomrule
\end{tabular}
}
\end{table}

\paragraph{Full results for TuneTables-medium and TuneTables-light}
Recall at the end of \cref{sec:experiments}, we introduced TuneTables-medium and TuneTables-light, showing that they can substantially decrease the runtime of the standard TuneTables with just a small reduction in accuracy.
While we presented summary results in \cref{tab:runtimes_summary}, now we present the full results in \cref{tab:runtimes_full}.
For descriptions of TuneTables variants, see \cref{sec:experiments}.

\begin{table}[t]
\caption{
\textbf{TuneTables-medium and TuneTables-light are substantially faster with only a modest decrease in accuracy.}
    We compare the average accuracy and runtime in seconds of three versions of TuneTables, across all 19 datasets of size $<50$K, as well as all 29 datasets.
    All datasets above the line were included in \cref{tab:less_than_50k}.
\label{tab:runtimes_full}
}
\resizebox{\textwidth}{!}{%
\centering
\begin{tabular}{lrrrrrrrrr}
\toprule
Dataset & \multicolumn{2}{c}{TuneTables-standard} & \multicolumn{2}{c}{TuneTables-medium} & \multicolumn{2}{c}{TuneTables-light} \\
 & Acc.\ & Runtime & Acc.\ & Runtime & Acc.\ & Runtime \\
\midrule
breast-cancer & 0.770 & 65 & 0.782 & 36 & 0.747 & 30 \\
heart-c & 0.903 & 66 & 0.892 & 27 & 0.892 & 23 \\
ecoli & 0.843 & 66 & 0.833 & 36 & 0.833 & 30 \\
colic & 0.892 & 66 & 0.892 & 27 & 0.892 & 23 \\
dresses-sales & 0.580 & 122 & 0.600 & 81 & 0.600 & 37 \\
climate & 0.951 & 97 & 0.963 & 38 & 0.963 & 30 \\
cylinder-bands & 0.846 & 82 & 0.870 & 41 & 0.870 & 34 \\
balance-scale & 0.995 & 55 & 0.995 & 26 & 0.995 & 22 \\
blood-transfusion & 0.782 & 56 & 0.773 & 31 & 0.760 & 30 \\
cmc & 0.556 & 109 & 0.536 & 32 & 0.514 & 44 \\
kc-1 & 0.856 & 187 & 0.843 & 68 & 0.842 & 50 \\
bioresponse & 0.798 & 3012 & 0.776 & 2374 & 0.766 & 653 \\
christine & 0.755 & 3920 & 0.740 & 3254 & 0.724 & 1366 \\
dilbert & 0.992 & 3749 & 0.990 & 2988 & 0.896 & 3107 \\
robert & 0.414 & 2397 & 0.420 & 1433 & 0.313 & 767 \\
har & 0.981 & 2657 & 0.985 & 2287 & 0.963 & 568 \\
eeg-eye-state & 0.986 & 1929 & 0.978 & 1752 & 0.938 & 213 \\
elevators & 0.900 & 1297 & 0.898 & 1110 & 0.897 & 211 \\
riccardo & 0.995 & 5247 & 0.995 & 3852 & 0.987 & 1309 \\
\midrule

volkert & 0.693 & 6331 & 0.707 & 5872 & 0.562 & 1116 \\
connect-4 & 0.817 & 5395 & 0.813 & 4699 & 0.682 & 471 \\
higgs & 0.714 & 4084 & 0.714 & 3153 & 0.648 & 719 \\
BNG (vote) & 0.974 & 2493 & 0.973 & 517 & 0.970 & 113 \\
albert & 0.658 & 17518 & 0.654 & 2056 & 0.642 & 350 \\
airlines & 0.653 & 44434 & 0.643 & 2230 & 0.601 & 352 \\
BNG (labor) & 0.967 & 7717 & 0.963 & 3942 & 0.956 & 641 \\
agrawal1 & 0.950 & 45504 & 0.950 & 2131 & 0.949 & 531 \\
poker-hand & 1.000 & 10471 & 0.998 & 8184 & 0.575 & 630 \\
click-prediction-small & 0.837 & 33148 & 0.835 & 3050 & 0.834 & 635 \\
\midrule
Average &  0.830  &	6975  &	0.828  &	1907	&  0.787  &	486 \\
\bottomrule
\end{tabular}
}
\end{table}

\subsection{Ablation study}
We report our results ablating the core components of our methods: the prompt length, its use during training and inference, the use of ensembles, and the prompt tuning procedure itself (against regular finetuning). Beginning with the prompt length, we see in \cref{tab:ablation} that while some dataset results are not sensitive to the prompt length, others vary significantly, and generally enjoy longer prompts. 
Ablating the importance of having real-data context during inference (same table) we find that it is important for smaller datasets. On the larger datasets, the variant with no real data at all is better in specific cases. However, having no such data during inference, a model would perform better not having it during training as well. Introducing real data first during inference is usually harmless, but significantly deteriorates the results on specific cases. Fine tuning the entire model is not only significantly more memory-intensive but also underperforms TuneTables in terms of accuracy.

In \cref{tab:ensembling-ablation} we show the results of our method with or without ensembles, and with or without context (in training and inference). Our ensembling strategy is beneficial in both cases. When using ensembles, the results with or without context are generally similar, although one variant may outperform the other on specific datasets.

%\begin{comment}
\begin{table}[t]
\centering
\caption{\textbf{Ablation on TuneTables variants.} Comparison of the different variants of our method (TabPFN-PT) and the entire backbone fine-tuning (TabPFN-FT) to TuneTables itself on the \newbenchmark benchmark. TuneTables is our full method described in \cref{sec:tunetables}. For each variant, we show its test accuracy. For all methods in this table except for TuneTables, there is no hyperparameter optimization (HPO). For TuneTables, we use our standard grid search.}

\label{tab:ablation}
\resizebox{\columnwidth}{!}{%
\begin{tabular}{lrrrrrrrr}
\toprule
 & TabPFN-FT & TabPFN-PT & TabPFN-PT & TabPFN-PT & TabPFN-PT & TabPFN-PT & TabPFN-PT & TuneTables \\
\textbf{Context in Training} & \cmark &  &  & \cmark & \cmark & \cmark & \cmark & Varied \\
\textbf{Context in Inference}  & \cmark & \cmark & & \cmark &  & \cmark & \cmark & Varied \\
\textbf{Tuned prompt length} & N/A & 1000 & 1000 & 1000 & 1000 & 100 & 10 & Varied \\
\midrule
Agrawal1 & 0.946 & 0.948 & 0.950 & 0.950 & 0.946 & 0.950 & 0.949 & 0.950 \\
BNG(labor) & 0.931 & 0.961 & 0.965 & 0.965 & 0.966 & 0.965 & 0.960 & 0.967 \\
BNG(vote) & 0.964 & 0.971 & 0.974 & 0.973 & 0.973 & 0.974 & 0.971 & 0.974 \\
Bioresponse & 0.754 & 0.756 & 0.760 & 0.747 & 0.668 & 0.729 & 0.748 & 0.798 \\
Click\_prediction\_small & 0.834 & 0.834 & 0.834 & 0.835 & 0.835 & 0.835 & 0.833 & 0.837 \\
airlines & 0.588 & 0.628 & 0.645 & 0.643 & 0.639 & 0.640 & 0.628 & 0.653 \\
albert & 0.605 & 0.646 & 0.648 & 0.651 & 0.651 & 0.657 & 0.648 & 0.658 \\
balance-scale & 0.937 & 0.909 & 0.944 & 0.940 & 0.456 & 0.932 & 0.933 & 0.995 \\
blood-transfusion %-service-center 
& 0.787 & 0.760 & 0.785 & 0.764 & 0.691 & 0.773 & 0.771 & 0.782 \\
breast-cancer & 0.730 & 0.743 & 0.713 & 0.741 & 0.661 & 0.724 & 0.718 & 0.770 \\
christine & 0.711 & 0.723 & 0.716 & 0.714 & 0.685 & 0.712 & 0.700 & 0.755 \\
climate-model %-simulation-crashes 
& 0.947 & 0.938 & 0.938 & 0.941 & 0.666 & 0.923 & 0.938 & 0.951 \\
cmc & 0.542 & 0.525 & 0.559 & 0.540 & 0.436 & 0.532 & 0.548 & 0.556 \\
colic & 0.888 & 0.871 & 0.880 & 0.816 & 0.563 & 0.870 & 0.842 & 0.892 \\
connect-4 & 0.678 & 0.787 & 0.810 & 0.795 & 0.804 & 0.767 & 0.679 & 0.817 \\
cylinder-bands & 0.842 & 0.791 & 0.759 & 0.870 & 0.556 & 0.820 & 0.856 & 0.846 \\
dilbert & 0.825 & 0.934 & 0.941 & 0.895 & 0.847 & 0.869 & 0.834 & 0.992 \\
dresses-sales & 0.550 & 0.573 & 0.533 & 0.567 & 0.553 & 0.560 & 0.577 & 0.580 \\
ecoli & 0.814 & 0.765 & 0.775 & 0.823 & 0.314 & 0.833 & 0.823 & 0.843 \\
eeg-eye-state & 0.579 & 0.626 & 0.651 & 0.651 & 0.654 & 0.616 & 0.601 & 0.986 \\
elevators & 0.894 & 0.895 & 0.899 & 0.896 & 0.854 & 0.896 & 0.892 & 0.902 \\
har & 0.940 & 0.980 & 0.982 & 0.971 & 0.946 & 0.962 & 0.945 & 0.981 \\
heart-c & 0.863 & 0.839 & 0.860 & 0.806 & 0.718 & 0.838 & 0.855 & 0.903 \\
higgs & 0.638 & 0.700 & 0.703 & 0.696 & 0.700 & 0.698 & 0.675 & 0.714 \\
kc1 & 0.843 & 0.848 & 0.843 & 0.853 & 0.218 & 0.860 & 0.851 & 0.856 \\
poker-hand & 0.601 & 0.479 & 0.993 & 0.982 & 0.985 & 0.690 & 0.572 & 1.000 \\
riccardo & 0.918 & 0.990 & 0.993 & 0.990 & 0.990 & 0.960 & 0.936 & 0.995 \\
robert & 0.303 & 0.394 & 0.404 & 0.386 & 0.369 & 0.354 & 0.311 & 0.414 \\
volkert & 0.487 & 0.621 & 0.647 & 0.631 & 0.649 & 0.605 & 0.540 & 0.693 \\
\midrule
Average  & 0.757  &	0.774  &	0.797	& 
 0.794  &	0.689  &	0.777  &	0.763 
 &	0.830 \\
\bottomrule
\end{tabular}
}
\end{table}
%\end{comment}

\section{TuneTables additional details}
\label{app:pt-experiments}

\paragraph{Tuned prompt implementation details. } We implement our tuned prompts by prepending randomly initialized vectors of identical dimension to the TabPFN encoder to the real data points fed to TabPFN as context during training. 

In the \textbf{CT} (`context in training') setting, we continue to provide real data points as part of the training context; the quantity of real data provided for each batch is drawn with uniform probability from a random variable which ranges from zero to a fixed upper bound, which is passed to the model as a hyperparameter. Usually that upper bound is $1152$, the default setting for TabPFN, unless the dataset is extremely small, in which case we select it according to the limitations of that dataset.

In the \textbf{NCT} (`no context in training') setting, no real data is provided as part of the training context; all data in the training set is used to fit the randomly initialized prompt. There are always a fixed number of data points provided, and it is the same for every batch. Usually that number is $128$, unless the dataset is extremely small, in which case we select it according to the limitations of that dataset.

We note that it is also possible to evaluate models trained in either setting either with context (C) or without (NC). In our experiments, we always evaluate all models on both, and report the setting with higher validation accuracy at test time. 

We find that on smaller datasets, the setting with context (C) sometimes outperforms the setting without context (NC), even after the prompt has been fitted, perhaps because the tuned prompt overfits on a small amount of available training data. However, when using ensembling, for datasets with more than 3000 samples in the training set, the NC setting is as good or better than the C setting. See  \cref{tab:ensembling-ablation}. We leave further investigation of this phenomenon to future work.

\paragraph{Loss function.} 

Most TuneTables experiments are optimized using the cross-entropy loss between the labels assigned to the training data (P) and TuneTables's outputs (Q):

\begin{equation}
    H(P, Q) = -\sum_{x} P(x) \log Q(x)
\end{equation}

To reduce the effects of overfitting on datasets with fewer samples, we extend our grid search to include the choice of loss function. We find that many such datasets benefit from fitting the tuned prompt via the KL divergence loss between the PFN's outputs (P) and TuneTables (Q).

\begin{equation}
    KL(P||Q) = \sum_{x} P(x) \log \left(\frac{P(x)}{Q(x)}\right)
\end{equation}

\paragraph{Compute.} We conduct all of our prompt tuning experiments on Google Cloud Platform (GCP), using a single NVIDIA L4 TPU with 24GB VRAM for each experiment.

\paragraph{Size of tuned prompts. } We ablate the size of tuned prompts in \cref{tab:ablation} and find that the larger tuned prompts generally perform better on datasets with more samples, while smaller tuned prompts excel on datasets with few samples. For this reason, the experimental results in \cref{sec:experiments} are reported on prompts of size 10 or 1000, conditioned on the number of samples in the input dataset. 

\paragraph{Duration of training. } Our experiments run for up to 100 epochs. We employ early stopping if our key metric (Accuracy) fails to improve after a fixed number of epochs. 
%(3 and 10, respectively) 

\paragraph{Evaluation of tuned prompts. } We validate our tuned prompts every epoch on a subset of the entire validation set if the validation set is large. At a fixed interval, we run the entire validation and test sets. After the experiment concludes, we report results from the epoch with the best key metric score.

\paragraph{Fine-tuned setting. } In the fine-tuned setting, our parameters are the same, except that we use a lower fixed learning rate of $1e-3$.

\paragraph{Ensembling over tuned prompts. } While individual tuned prompts are already a substantial improvement on TabPFN for sample-large datasets, we find that these improvements sometimes compound when we ensemble over multiple tuned prompts.

We draw the inspiration for our ensembling approach from the internal ensembling used in TabPFN, which averages predictions over permutations of feature indices and label indices\citep{hollmann2022tabpfn}. For the results presented in \cref{sec:experiments}, we ensemble over ten permutations of each dataset, averaging the top two ensemble member predictions (as measured by NC accuracy on the validation set in the NCT setting, or C accuracy in the CT setting).

In a TuneTables ensemble, each ensemble member fits its own tuned prompt to the data. Variance in the ensemble members is introduced by differences in the random initialization of the tuned prompt, as well as permuting the order of features and labels, a la TabPFN, one time before each tuned prompt is fitted. 

See \cref{tab:ensembling-ablation} for ablation studies of the effectiveness of ensembling.

\begin{table*}[t]
\caption{\textbf{Ensembled models outperform models trained on a single tuned prompt; with ensembling, the training and testing without real-data context (NC) setting matches or exceeds the setting with context (C).} Runs with prompt tuned only once are noted as TabPFN-PT, and ensembles of such runs as TabPFN-PT-Ens. Although the improvements are often quite small, we find that ensembles generally outperform single tuned prompts in both C and NC settings. We also find that ensembles trained without additional real data context at train time or test time are as good or better than ensembles trained and tested with real data context on datasets larger than 3000 samples. 
}
\label{tab:ensembling-ablation}
\resizebox{\textwidth}{!}{%
\centering
\begin{tabular}{lrrrrr}
\toprule
Dataset & TabPFN-PT-C & TabPFN-PT-Ens-C & TabPFN-PT-NC & TabPFN-PT-Ens-NC \\
\midrule
agrawal1 & 0.949 & \textbf{0.95} & \textbf{0.95} & 0.949 \\
airlines & 0.645 & \textbf{0.649} & 0.645 & 0.646 \\
albert & 0.657 & \textbf{0.66} & 0.648 & \textbf{0.66} \\
balance-scale & 0.921 & 0.968 & \textbf{0.984} & 0.952 \\
bioresponse & 0.763 & \textbf{0.795} & 0.776 & 0.776 \\
blood-transfusion & 0.813 & \textbf{0.84} & 0.827 & 0.747 \\
BNG (labor) & 0.965 & 0.966 & 0.965 & \textbf{0.967} \\
BNG (vote) & 0.974 & 0.976 & 0.975 & \textbf{0.977} \\
breast-cancer & \textbf{0.793} & \textbf{0.793} & 0.759 & 0.69 \\
car & 0.96 & 0.971 & \textbf{0.977} & 0.965 \\
christine & \textbf{0.76} & 0.738 & 0.734 & 0.756 \\
click-prediction-small & 0.834 & \textbf{0.836} & 0.834 & \textbf{0.836} \\
climate & 0.926 & 0.944 & 0.963 & \textbf{0.981} \\
cmc & 0.534 & \textbf{0.581} & 0.547 & 0.446 \\
colic & 0.811 & \textbf{0.892} & 0.865 & 0.865 \\
connect-4 & 0.796 & \textbf{0.814} & 0.808 & 0.812 \\
cylinder-bands & 0.815 & 0.815 & \textbf{0.926} & 0.778 \\
dilbert & 0.87 & 0.951 & 0.948 & \textbf{0.968} \\
dresses-sales & 0.6 & \textbf{0.68} & 0.66 & 0.6 \\
eeg-eye-state & 0.74 & 0.977 & 0.666 & \textbf{0.983} \\
elevators & 0.903 & \textbf{0.908} & 0.902 & 0.902 \\
har & 0.94 & 0.984 & 0.976 & \textbf{0.987} \\
heart-c & 0.871 & \textbf{0.903} & 0.871 & 0.871 \\
higgs & 0.695 & \textbf{0.712} & 0.709 & 0.709 \\
kc-1 & 0.867 & \textbf{0.872} & 0.867 & \textbf{0.872} \\
poker-hand & \textbf{1} & \textbf{1} & 0.992 & \textbf{1} \\
riccardo & 0.991 & \textbf{0.996} & 0.994 & \textbf{0.996} \\
robert & 0.391 & 0.415 & 0.421 & \textbf{0.444} \\
volkert & 0.633 & 0.665 & 0.658 & \textbf{0.672} \\
\midrule
Average & 0.807 &	0.836 &	0.822  &	0.821 \\
\bottomrule
\end{tabular}
}
\end{table*}

\section{Regression experiments}

In this section, we show that using prompt tuning, we are also able to extend PFNs to perform well on regression datasets, further highlighting the flexibility of our approach.

\begin{table}[]
\centering
\caption{\textbf{TuneTables is competitive with strong baselines on a range of regression datasets.} In this table, we compare 7 algorithms on \nregdatasets regression datasets from \cite{mcelfresh2023neural}. The terminology and methods of this table follow \cref{tab:less_than_50k}, except that we report mean $R^2$ score instead of mean accuracy as our primary statistic. TuneTables ties with XGBoost for the number of wins, has the highest average $R^2$ score of any algorithm, and the second-highest mean rank after LightGBM.}
\label{tab:ttregzscore}
\resizebox{\columnwidth}{!}{%
\begin{tabular}{lllllll}
\toprule
Model & Mean R2 & Mean Rank & Mean Z-Scores & Std Z-Scores & Med Z-Scores & Number of Wins \\ \midrule
TuneTables & 0.603 & 3.533 & 0.792 & 0.448 & 0.768 & 4.0 \\
LightGBM & 0.536 & 3.333 & 0.640 & 0.472 & 0.660 & 1.0 \\
CatBoost & 0.508 & 3.800 & 0.431 & 0.933 & 0.576 & 2.0 \\
XGBoost & 0.386 & 4.000 & 0.116 & 1.009 & 0.405 & 3.0 \\
MLP & 0.359 & 5.267 & -0.009 & 0.673 & 0.109 & 1.0 \\
STG & 0.332 & 6.067 & -0.369 & 0.864 & -0.169 & 2.0 \\
TabNet & 0.239 & 5.867 & -0.107 & 0.747 & 0.099 & 0.0 \\
VIME & -0.007 & 7.733 & -1.072 & 1.054 & -1.159 & 0.0 \\
DeepFM & -7.954 & 5.400 & -0.422 & 1.248 & 0.027 & 2.0 \\
\bottomrule
\end{tabular}%
}
\end{table}

\subsection{Adapting TuneTables to regression problems.} Adapting TuneTables to regression problems introduces new challenges, as TabPFN was not designed for use with such problems. In particular, we find that end-to-end fine-tuning outperforms prompt tuning in this setting, and that the most effective grid search for regression problems utilizes a space of \textit{PFN base models}. Put another way, TuneTables-regression does not use prompt tuning, and does not always use TabPFN as its foundation model. Rather, it searches a space of foundation models to find the best performer for a particular regression problem. We search over three foundation models; TabPFN, the checkpoint released in \cite{muller2022transformers} and a new PFN we train from scratch for 10 epochs on a synthetic prior of regression-specific datasets.

\subsection{Details on the datasets used for the experiments and baseline experimental design}

Similar to \cite{mcelfresh2023neural}, each algorithm is tuned for each dataset
by maximizing the R-squared (R2) metric. Each dataset corresponds to an OpenML task, and can be preprocessed exactly like the classification datasets used in other experiments. The \nregdatasets~ datasets used in these experiments are “Bank-Note-Authentication-
UCI” (OpenML task 361002), “EgyptianSkulls” (5040), “Wine” (190420), “Wisconsin-breast-
cancer-cytology-features” (361003), “bodyfat” (5514), “california” (361089), “chscase-foot” (5012), “cleveland” (2285), “colleges” (359942), “cpu-small” (4883), “liver-disorders” (52948), “meta” (4729), “mv” (4774), “pbc” (4850), and “veteran” (4828). Unlike  prior work, we report non-normalized averages, as we believe it gives a more realistic indication of real-world performance at a glance.

We use 12 deep learning and boosted tree algorithms as baselines; VIME, TabTransformer, TabNet, DANet, STG, MLP, LightGBM, CatBoost, XGBoost, NODE, DeepFM.

\begin{figure*}[t]
    \centering
    \includegraphics[width=0.7\textwidth]{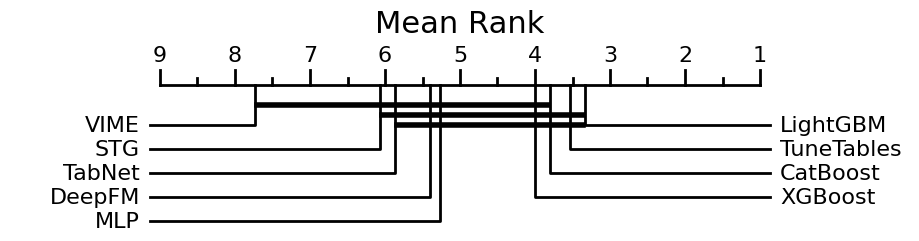}
    \caption{\textbf{TuneTables is competitive with state-of-the-art tabular models on regression.} 
    }
    \label{fig:critical_difference_regression}
\end{figure*}

\subsection{Results of regression experiments}

In \cref{tab:ttregzscore} results and \cref{fig:critical_difference_regression}, we show the results of our experiments. Overall, we see that TuneTables is competitive with top-performing algorithms across a variety of metrics; mean win rank, average $R^2$ score, etc. Equally importantly, TuneTables is significantly stronger on several datasets than both boosted trees and deep baselines.
\section{Summarization details and decision boundaries of prompt-tuned TabPFN on toy 2D Problems} \label{app:interpretability}

\begin{figure}
    \centering
    \includegraphics[width=\textwidth]{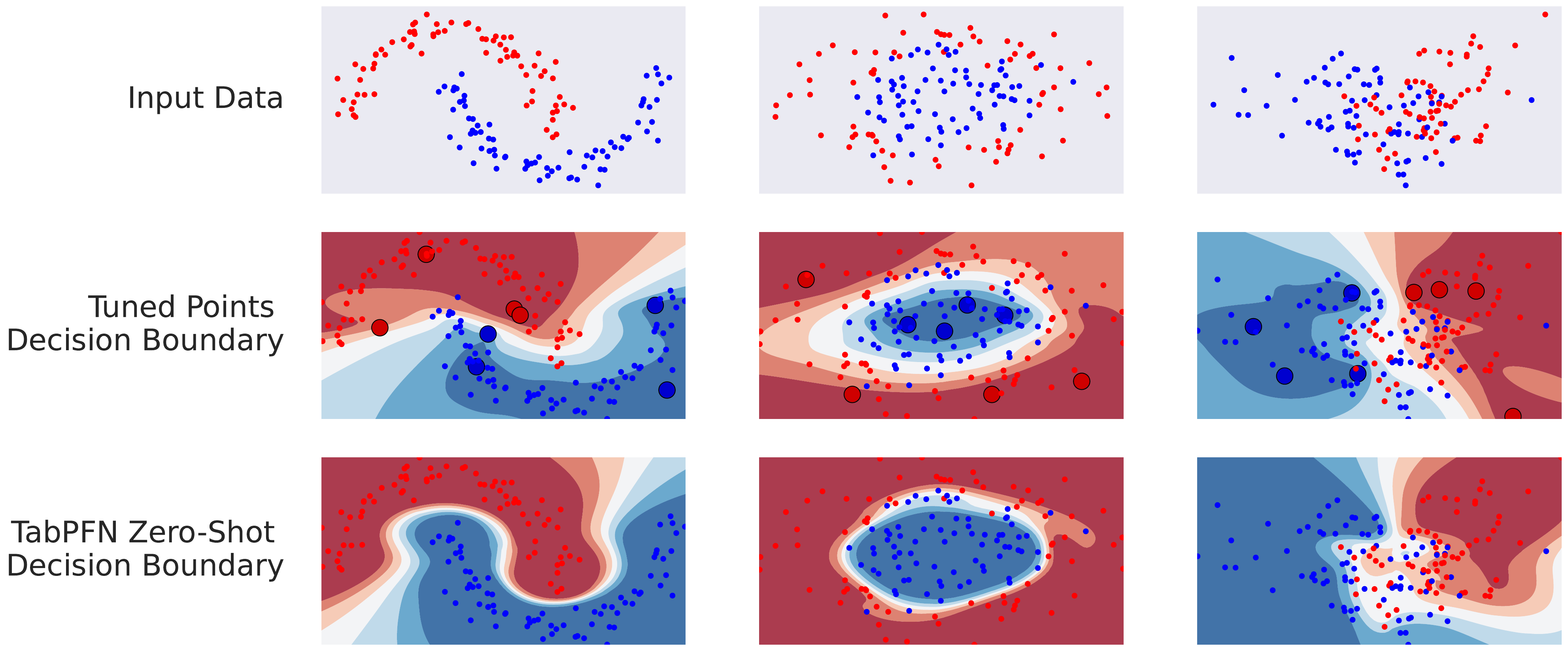}
    \caption{\textbf{Decision boundaries on 2D datasets.} Shown are prompt-tuned TabPFN decision boundaries as well as the regular zero-shot TabPFN decision boundaries. Larger dots in middle row represent the tuned points for the two classes.
    }
    \label{fig:contours}
\end{figure}

For the experiments with prompts of only two examples, we prompt-tune one example per class for 500 epochs. The prompt is input directly into TabPFN without undergoing any preprocessing steps that the original data went through. Therefore, the tuned feature values do not correspond directly to the original values in the dataset. For the breast cancer dataset, it reaches 100\% accuracy; for the diabetes dataset, it reaches 78.7\% accuracy. 

We also show decision boundaries of TabPFN and prompt-tuned TabPFN on toy 2d classification problems using scikit-learn \citep{pedregosa2011scikit} in Fig.\ \ref{fig:contours}. The prompt contains four datapoints for each class. One can see how the points are tuned to recover the decision boundaries. 
Due to the very low prompt length and the low dataset size, the prompt-tuned decision boundaries are slightly less accurate than TabPFN (consistent with our results for small datasets in \cref{tab:ablation}). 

\begin{figure}
    \centering
    \includegraphics[width=.4\textwidth]{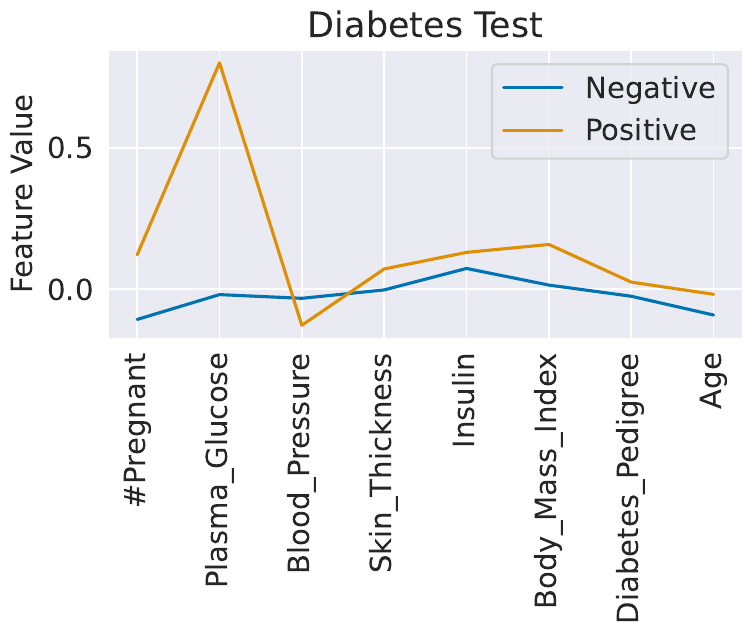}
    \vspace{-0.20cm}
    \caption{\textbf{Diabetes dataset \citep{smith1988using} with high accuracies from just two samples.}}
    \label{fig:breast_cancer_rest}
\end{figure}

\section{Results for Deeper GBDTs} 
\label{app:deepgbdt}

In the results in \cref{sec:experiments}, we compare TuneTables to the GBDTs, XGBoost and CatBoost, using a search space in which the number of trees can range from 1 to 1000.
Now, we compare TuneTables to XGBoost and CatBoost when they have expanded search spaces, where the number of trees can range from 1 to 10\,000.
We re-compute the critical difference plot and table results:
see \cref{fig:critical_difference_gdbt_var}, \cref{tab:less_than_50k_gdbt_var}, and \cref{tab:full_results_gbdt_var}.

In \cref{fig:critical_difference_gdbt_var} and \cref{tab:less_than_50k_gdbt_var}, we see that TuneTables still performs favorably compared to CatBoost and XGBoost on datasets of size up to 50\,000.
Note that the larger search space may perform better on larger datasets, it is also more challenging to search through in 30 iterations.
We present the full results on all datasets in \cref{tab:full_results_gbdt_var}.

\begin{table*}[!htb]
\caption{\textbf{TuneTables matches SOTA algorithms on small and medium-sized datasets.}
In this table, we compare algorithms over 19 datasets in \newbenchmark with at most 50\,000 samples. For each algorithm, we compute its mean accuracy, mean runtime, and mean rank in terms of accuracy.
We also compute the mean Z-score, computed by normalizing the set of results on each dataset (by mean 0 std.\ 1), so that each dataset has the same weight, and averaging each algorithm's normalized performances. Std.\ Z-Score is computed with respect to random splits and averaged across datasets. Fractional num.\ wins values are averaged over three splits per dataset, and reflect the presence of multi-way ties on certain splits.
This table is similar to \cref{tab:less_than_50k}, but the benchmark is \newbenchmark, and the search spaces for XGBoost and CatBoost are expanded to include more trees.
}
\label{tab:less_than_50k_gdbt_var}
\resizebox{\textwidth}{!}{%
\centering
\begin{tabular}{lrrrrrrr}
\toprule
Method & Mean Acc.\  & Mean Rank & Mean Z-Score & Std.\ Z-Score & Med.\ Z-Score & Num.\ Wins & Runtime \\
\midrule
TuneTables & 0.831 & 2.263 & 0.471 & 0.536 & 0.614 & 9.833 & 1325 \\
CatBoost & 0.827 & 2.579 & 0.320 & 0.644 & 0.558 & 4.500 & 915\\
XGBoost & 0.825 & 2.684 & 0.267 & 0.625 & 0.720 & 3.500 & 304 \\
rtdl\_ResNet & 0.790 & 3.561 & -0.341 & 0.461 & -0.408 & 0.333  & 500 \\
rtdl\_MLP & 0.755 & 3.912 & -0.717 & 0.678 & -0.897 & 0.833 & 374 \\
\bottomrule
\end{tabular}
}
\end{table*}

\begin{table}[t]
\caption{\textbf{Comparison of top-performing methods on the \newbenchmark benchmark.} For each algorithm, we show its test accuracy and runtime (seconds).
We also show the number of samples in each dataset.
All datasets above the line were included in \cref{tab:less_than_50k_gdbt_var}.
This table is similar to \cref{tab:full_results}, but the search spaces for XGBoost and CatBoost are expanded to include more trees.
}
\label{tab:full_results_gbdt_var}
\resizebox{\textwidth}{!}{%
\centering
\begin{tabular}{lrrrrrrrrr}
\toprule
Dataset & Size & \multicolumn{2}{c}{TabPFN} & \multicolumn{2}{c}{TuneTables} & \multicolumn{2}{c}{CatBoost} & \multicolumn{2}{c}{XGBoost} \\
& & Acc.\ & Runtime & Acc.\ & Runtime  & Acc.\ & Runtime  & Acc.\ & Runtime \\
\midrule
breast-cancer & 286 & 0.765 & 29 & 0.770 & 65 & 0.736 & 190 & 0.724 & 23 \\
heart-c & 303 & 0.848 & 40 & 0.903 & 66 & 0.839 & 164 & 0.849 & 27 \\
ecoli & 336 & 0.848 & 30 & 0.843 & 66 & 0.892 & 394 & 0.843 & 72 \\
colic & 368 & 0.856 & 39 & 0.892 & 66 & 0.874 & 458 & 0.883 & 27 \\
dresses-sales & 500 & 0.578 & 41 & 0.580 & 122 & 0.613 & 292 & 0.660 & 30 \\
cylinder-bands & 540 & 0.800 & 41 & 0.846 & 82 & 0.827 & 552 & 0.846 & 35 \\
climate & 540 & 0.959 & 59 & 0.951 & 97 & 0.951 & 282 & 0.938 & 29 \\
balance-scale & 625 & 0.990 & 29 & 0.995 & 55 & 0.931 & 241 & 0.894 & 112 \\
blood-transfusion & 748 & 0.801 & 25 & 0.782 & 56 & 0.756 & 92 & 0.760 & 27 \\
cmc & 1473 & 0.554 & 91 & 0.556 & 109 & 0.547 & 224 & 0.552 & 58 \\
kc-1 & 2109 & 0.862 & 168 & 0.856 & 187 & 0.855 & 138 & 0.853 & 40 \\
bioresponse & 3151 & 0.797 & 638 & 0.798 & 3012 & 0.791 & 618 & 0.797 & 131 \\
christine & 5418 & 0.742 & 666 & 0.755 & 3920 & 0.735 & 234 & 0.742 & 158 \\
robert & 10000 & 0.250 & 964 & 0.414 & 2397 & 0.534 & 2934 & 0.524 & 3294 \\
dilbert & 10000 & 0.922 & 761 & 0.992 & 3749 & 0.987 & 3185 & 0.986 & 779 \\
har & 10299 & 0.936 & 370 & 0.981 & 2657 & 0.993 & 3004 & 0.995 & 251 \\
eeg-eye-state & 14980 & 0.940 & 178 & 0.986 & 1929 & 0.957 & 2594 & 0.955 & 116 \\
elevators & 16599 & 0.902 & 186 & 0.902 & 1297 & 0.894 & 1316 & 0.896 & 103 \\
riccardo & 20000 & 0.922 & 1395 & 0.995 & 5247 & 0.996 & 473 & 0.996 & 468 \\

\toprule

volkert & 58310 & 0.567 & 459 & 0.693 & 6331 & 0.715 & 2364 & 0.711 & 1550 \\
higgs & 67557 & 0.671 & 931 & 0.714 & 4084 & 0.729 & 222 & 0.726 & 128 \\
connect-4 & 98050 & 0.668 & 696 & 0.817 & 5395 & 0.864 & 3492 & 0.860 & 593 \\
BNG (vote) & 131072 & 0.968 & 1976 & 0.974 & 2493 & 0.975 & 42 & 0.975 & 76 \\
albert & 425240 & 0.642 & 2363 & 0.658 & 17518 & 0.709 & 4001 & 0.694 & 529 \\
airlines & 539383 & 0.600 & 2602 & 0.653 & 44434 & 0.671 & 2023 & 0.673 & 305 \\
BNG (labor) & 1000000 & 0.937 & 5518 & 0.967 & 7717 & 0.971 & 226 & 0.970 & 184 \\
agrawal1 & 1000000 & 0.948 & 5158 & 0.950 & 45504 & 0.951 & 142 & 0.951 & 98 \\
poker-hand & 1025009 & 0.531 & 2423 & 1.000 & 10471 & 1.000 & 6039 & 0.997 & 3469 \\
click-prediction-small & 1997410 & 0.833 & 10421 & 0.837 & 33148 & 0.842 & 84 & 0.843 & 102 \\
\midrule
Average  & & 0.781 &	1321  &	0.830  &	6975  &	0.832  &	1242  &	0.831  &	703 \\
\bottomrule
\end{tabular}
}
\end{table}

\clearpage

\section*{NeurIPS Paper Checklist}

\begin{enumerate}

\item {\bf Claims}
    \item[] Question: Do the main claims made in the abstract and introduction accurately reflect the paper's contributions and scope?
    \item[] Answer: \answerYes{} % Replace by \answerYes{}, \answerNo{}, or \answerNA{}.
    \item[] Justification: Our abstract and introduction both state the main claims in the paper.

\item {\bf Limitations}
    \item[] Question: Does the paper discuss the limitations of the work performed by the authors?
    \item[] Answer: \answerYes{} % Replace by \answerYes{}, \answerNo{}, or \answerNA{}.
    \item[] Justification: We discuss the limitations of our work in \cref{sec:conclusion}.

\item {\bf Theory Assumptions and Proofs}
    \item[] Question: For each theoretical result, does the paper provide the full set of assumptions and a complete (and correct) proof?
    \item[] Answer: \answerNA{} % Replace by \answerYes{}, \answerNo{}, or \answerNA{}.
    \item[] Justification: Our work does not include theoretical results.

    \item {\bf Experimental Result Reproducibility}
    \item[] Question: Does the paper fully disclose all the information needed to reproduce the main experimental results of the paper to the extent that it affects the main claims and/or conclusions of the paper (regardless of whether the code and data are provided or not)?
    \item[] Answer: \answerYes{} % Replace by \answerYes{}, \answerNo{}, or \answerNA{}.
    \item[] Justification: \cref{sec:experiments} and \cref{app:experiments} lay out all information needed to reproduce the experimental results.

\item {\bf Open access to data and code}
    \item[] Question: Does the paper provide open access to the data and code, with sufficient instructions to faithfully reproduce the main experimental results, as described in supplemental material?
    \item[] Answer: \answerYes{} % Replace by \answerYes{}, \answerNo{}, or \answerNA{}.
    \item[] Justification: We do release all code and materials needed to reproduce our results. We provided this link in the introduction: \url{https://anonymous.4open.science/r/TuneTables}.

\item {\bf Experimental Setting/Details}
    \item[] Question: Does the paper specify all the training and test details (e.g., data splits, hyperparameters, how they were chosen, type of optimizer, etc.) necessary to understand the results?
    \item[] Answer: \answerYes{} % Replace by \answerYes{}, \answerNo{}, or \answerNA{}.
    \item[] Justification: \cref{sec:experiments} and \cref{app:experiments} specify all experimental settings and details.

\item {\bf Experiment Statistical Significance}
    \item[] Question: Does the paper report error bars suitably and correctly defined or other appropriate information about the statistical significance of the experiments?
    \item[] Answer: \answerYes{} % Replace by \answerYes{}, \answerNo{}, or \answerNA{}.
    \item[] Justification: Our results in \cref{sec:experiments} do include standard deviations, and we also include critical difference plots in order to compute the statistical significance of our results.

\item {\bf Experiments Compute Resources}
    \item[] Question: For each experiment, does the paper provide sufficient information on the computer resources (type of compute workers, memory, time of execution) needed to reproduce the experiments?
    \item[] Answer: \answerYes{} % Replace by \answerYes{}, \answerNo{}, or \answerNA{}.
    \item[] Justification: \cref{sec:experiments} does provide information on the compute resources needed to reproduce the experiments.
    
\item {\bf Code Of Ethics}
    \item[] Question: Does the research conducted in the paper conform, in every respect, with the NeurIPS Code of Ethics \url{https://neurips.cc/public/EthicsGuidelines}?
    \item[] Answer: \answerYes{} % Replace by \answerYes{}, \answerNo{}, or \answerNA{}.
    \item[] Justification: The paper conforms to the NeurIPS Code of Ethics.

\item {\bf Broader Impacts}
    \item[] Question: Does the paper discuss both potential positive societal impacts and negative societal impacts of the work performed?
    \item[] Answer: \answerYes{} % Replace by \answerYes{}, \answerNo{}, or \answerNA{}.
    \item[] Justification: We discuss the societal impacts in \cref{sec:impact}.
    
\item {\bf Safeguards}
    \item[] Question: Does the paper describe safeguards that have been put in place for responsible release of data or models that have a high risk for misuse (e.g., pretrained language models, image generators, or scraped datasets)?
    \item[] Answer: \answerNA{} % Replace by \answerYes{}, \answerNo{}, or \answerNA{}.
    \item[] Justification: As discussed in \cref{sec:impact}, our paper does not pose such risks.

\item {\bf Licenses for existing assets}
    \item[] Question: Are the creators or original owners of assets (e.g., code, data, models), used in the paper, properly credited and are the license and terms of use explicitly mentioned and properly respected?
    \item[] Answer: \answerYes{} % Replace by \answerYes{}, \answerNo{}, or \answerNA{}.
    \item[] Justification: We credit the original owners of the resources we use in the supplementary material.

\item {\bf New Assets}
    \item[] Question: Are new assets introduced in the paper well documented and is the documentation provided alongside the assets?
    \item[] Answer: \answerYes{} % Replace by \answerYes{}, \answerNo{}, or \answerNA{}.
    \item[] Justification: We release code which is documented with a readme and comments.

\item {\bf Crowdsourcing and Research with Human Subjects}
    \item[] Question: For crowdsourcing experiments and research with human subjects, does the paper include the full text of instructions given to participants and screenshots, if applicable, as well as details about compensation (if any)? 
    \item[] Answer: \answerNA{} % Replace by \answerYes{}, \answerNo{}, or \answerNA{}.
    \item[] Justification: Our paper does not involve crowdsourcing nor research with human subjects.

\item {\bf Institutional Review Board (IRB) Approvals or Equivalent for Research with Human Subjects}
    \item[] Question: Does the paper describe potential risks incurred by study participants, whether such risks were disclosed to the subjects, and whether Institutional Review Board (IRB) approvals (or an equivalent approval/review based on the requirements of your country or institution) were obtained?
    \item[] Answer: \answerNA{} % Replace by \answerYes{}, \answerNo{}, or \answerNA{}.
    \item[] Justification: Our paper does not involve crowdsourcing nor research with human subjects.

\end{enumerate}

\end{document}